\documentclass[runningheads]{llncs}

\usepackage{eccv}

\usepackage{eccvabbrv}

\usepackage{graphicx}
\usepackage{booktabs}
\usepackage{algorithm}
\usepackage{algpseudocode}
\usepackage{color,soul}
\usepackage{mdframed}
\usepackage{listings}
\usepackage{svg}

\definecolor{light-gray}{gray}{0.80}
\usepackage[accsupp]{axessibility}  

\usepackage{hyperref}

\usepackage{orcidlink}

\begin{document}

\title{BlenderAlchemy: Editing 3D Graphics with Vision-Language Models}

\titlerunning{BlenderAlchemy}

\author{Ian Huang \orcidlink{0009-0000-8426-7543} \and
Guandao Yang \orcidlink{0000-0002-2992-5803} \and
Leonidas Guibas \orcidlink{0000-0002-8315-4886}}

\authorrunning{Huang et al.}

\institute{Stanford University}

\maketitle

\begin{abstract}
Graphics design is important for various applications, including movie production and game design. To create a high-quality scene, designers usually need to spend hours in software like Blender, in which they might need to interleave and repeat operations, such as connecting material nodes, hundreds of times. Moreover, slightly different design goals may require completely different sequences, making automation difficult. In this paper, we propose a system that leverages Vision-Language Models (VLMs), like GPT-4V, to intelligently search the design action space to arrive at an answer that can satisfy a user's intent. Specifically, we design a vision-based edit generator and state evaluator to work together to find the correct sequence of actions to achieve the goal. Inspired by the role of visual imagination in the human design process, we supplement the visual reasoning capabilities of VLMs with ``imagined'' reference images from image-generation models, providing visual grounding of abstract language descriptions. In this paper, we provide empirical evidence suggesting our system can produce simple but tedious Blender editing sequences for tasks such as editing procedural materials and geometry from text and/or reference images, as well as adjusting lighting configurations for product renderings in complex scenes. \footnote{For project website and code, please go to: \url{https://ianhuang0630.github.io/BlenderAlchemyWeb/}}
\end{abstract}

\section{Introduction}
\label{sec:intro}

\begin{figure}[tb]
    \centering
    \includegraphics[width=\textwidth]{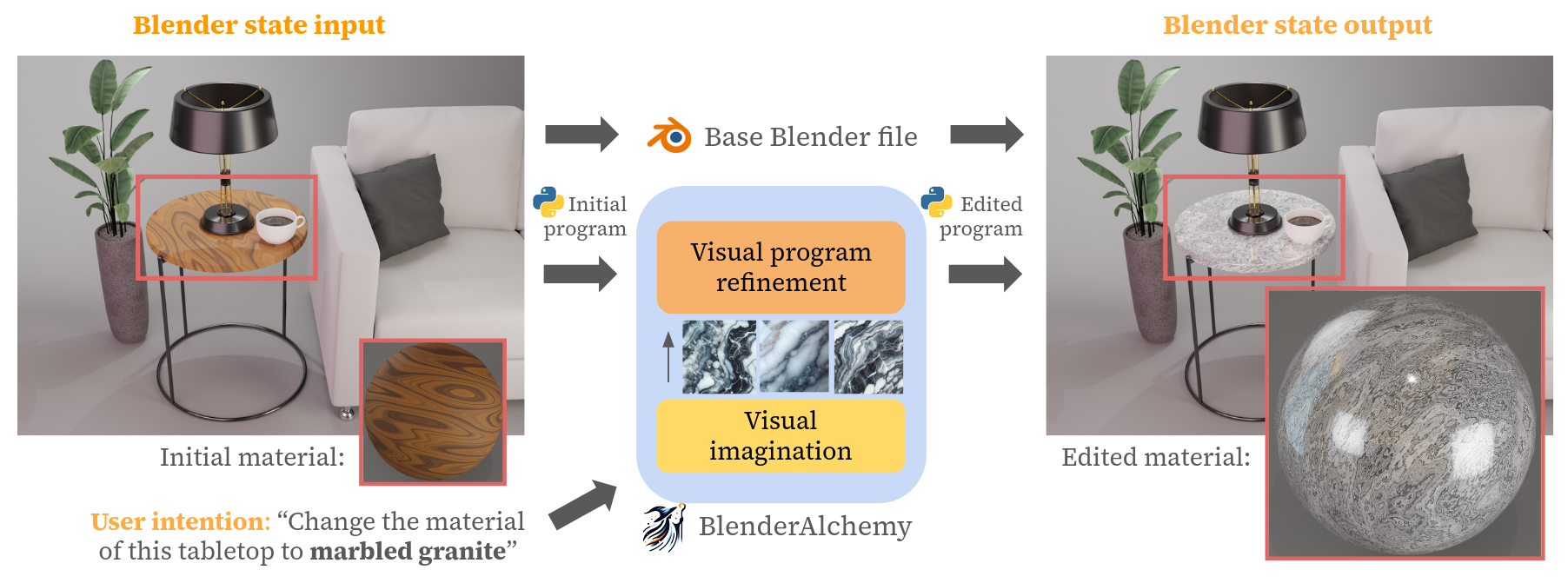}
    \vspace{-1em}
    \caption{\textbf{Overview of BlenderAlchemy.} Given an input Blender state and a user intention specified using either language or reference images,  BlenderAlchemy edits the Blender state to satisfy that intention by \setulcolor{orange} 
    \ul{\textit{iteratively refining a Blender python program}} that executes to produce the final Blender state. Our system additionally leverages text-to-image generation for  \setulcolor{yellow} \ul{\textit{visual imagination}}, a step that expands a text-only user intention to a concrete visual target to improve program refinement.}
    \label{fig:teaser}
    \vspace{-1em}
\end{figure}

To produce the compelling graphics content we see in movies or video games, 3D artists usually need to spend hours in software like Blender to find appropriate surface materials, object placements, and lighting arrangements.
These operations require the artist to create a mental picture of the target, experiment with different parameters, and visually examine whether their edits get closer to the end goal.
One can imagine automating these processes by converting language or visual descriptions of user intent into edits that achieve a design goal. Such a system can improve the productivity of millions of 3D designers and impact various industries that depend on 3D graphic design.


Graphic design is very challenging because even a small design goal requires performing a variety of different tasks. 
For instance, modeling of a game environment requires the 3D artist to cycle between performing modeling, material design, texture painting, animation, lighting, and scene composition.
Prior attempts usually focus on specific editing tasks, like material synthesis \cite{chen2023text2tex,zsolnai2018gaussian}.
While these approaches show promising performance in the tasks they are designed for, it is non-trivial to combine these task-specific methods to satisfy an intended design goal.
An alternative is to leverage Large Language Models (LLMs)~\cite{2023GPT4,touvron2023llama,jiang2023mistral} to digest user intent and suggest design actions by proposing intelligent combination of existing task-specific tools~\cite{schick2024toolformer} or predicting edits to programs step-by-step~\cite{wei2022chain,yao2024tree}.
While LLMs have excellent abilities to understand user intentions and suggest sequences of actions to satisfy them, applying LLMs to graphical design remains challenging largely because language cannot capture the visual consequences of actions performed in software like Blender.

One promising alternative is to leverage vision language models (VLM), such as LLaVA~\cite{li2024llava}, GPT-4V~\cite{2023GPT4VisionSC}, Gemini~\cite{team2023gemini}, and DallE-3~\cite{betker2023improving}.
These VLMs have shown to be highly capable of understanding detailed visual information \cite{yang2023dawnofllms,yin2023woodpecker,yin2023survey,fu2023mme} and generating compelling images~\cite{betker2023improving}. 
We posit that these VLMs can be leveraged to complete different kinds of design tasks within the Blender design environment, like editing materials, geometry and lighting setups.

In this paper, we provide a proof-of-concept system using the vision foundation model GPT-4V to generate and edit programs that modify the state of a Blender workspace to satisfy a user intention.
Specifically, we first initialize the state of a workspace within Blender.
The Blender state is parameterized as a short Python program, and a base Blender file.
The user will then input a text description and potentially a reference image to communicate the desirable design outcome.
The system is tasked to edit the program so that, when executed on the base Blender file, the rendered image can satisfy the user's intention.
Figure~\ref{fig:teaser} provides an illustration of the problem setup.

\begin{figure}[tb]
    \centering
    \includegraphics[width=\textwidth]{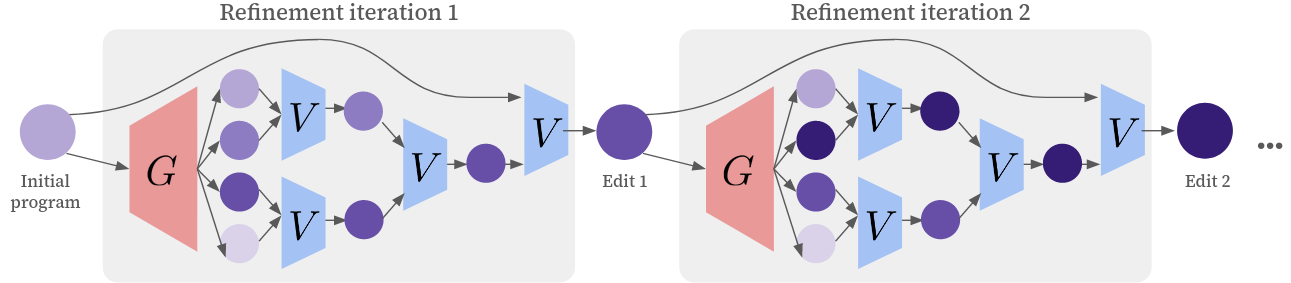}
    \vspace{-1em}
    \caption{\textbf{Iterative visual program editing} employs a edit generator $G$ and a state evaluator $V$ in each iteration to explore and prune different potential program edits, where $G$ generates plausible variants of an input program and $V$ picks between two programs based on the consequences they have to the Blender visual state and their alignment to the user intention. Each iteration of the refinement explores variations of the most promising program from the previous iteration. See Algorithm \ref{alg:iterative_refinement} for details.}
    \label{fig:iterative-visual-program-edit}
    \vspace{-1em}
\end{figure}

Naively applying VLMs to this editing setting gives rise to many failure cases, possibly due to the fact that out-of-the-box VLMs have a poor understanding of the visual consequences of Blender program edits.
To counter this, we propose a visually-guided program search procedure that combines a vision-aware edit generator and a visual state evaluator to iteratively search for a suitable program edit (Figure \ref{fig:iterative-visual-program-edit}).
Inspired by the human design process, our system performs guided trial-and-error, capped by some computation budget.
Within each iteration, the visual program generator will propose several possible edits on the current program.
These edits will be applied and executed in Blender to produce a rendered image.
These rendered images will be provided to visual evaluator, which will select the best renders via pairwise comparison by assessing which choice better satisfies the design goal specified by text and reference image.
The program achieving the best render will replace the current program as a starting point for the visual program generator.
This iterative search process, however, has very low success rate because of the sparsity of correct program edits in the vast program space.
To improve the success rate, we further propose two techniques.
First, when the proposed program of the new iteration does not contain a viable candidate, we revert to the best candidate of the prior program.
This reversion mechanism make sure the search procedure will not diverge when facing a batch of bad edit candidates.
Second, to facilitate our visual evaluator and generator to better understand user intent, we leverage the ``visual imagination'' of text-to-image generative model to imagine a reference image.
We show that our method is capable of accomplishing graphical design tasks within Blender, guided by user intention in the form of text and images. 
We demonstrate the effectiveness of our system on material, geometry and lighting design tasks, all parts of the 3D design process where artists spend a significant amount of time, ranging from 20 hrs to 4-6 workdays \textit{per} model~\cite{shahbazi2024,3dace}.
We show that our method can outperform prior works designed for similar problem settings, such as BlenderGPT~\cite{blendergpt}. In summary, the main contributions of this paper include:
\begin{enumerate}
    \item We propose BlenderAlchemy, a system that's able to edit visual programs based on user input in the form of text or images. 
    \item We identify key components that make the system work: a visual state evaluator, a visual edit generator, a search algorithm with an edit reversion mechanism, and a visual imagination module to facilitate the search. 
    \item We provide evidence showing that our system can outperform prior works in text-based procedural material editing, as well as its applicability to other design tasks including editing geometry and lighting configurations.
\end{enumerate}

\section{Related works}
\label{sec:related_works}

\vspace{-1em}\paragraph{\textbf{Task-specific tools for Material Design.}}
Large bodies of works have been dedicated to using learning-based approaches to generate materials. 
Prior works exploit 2D diffusion models to generate texture maps either in a zero-shot way \cite{chen2023text2tex,richardson2023texture,yang20233dstyle,huang2023aladdin,wen2023anyhome} or through fine-tuning \cite{sharma2023alchemist,zeng2023paint3d}.
Though these works open the possibility of generating and modifying textures of objects using natural language for 3D meshes, these works do not model the material properties in a way that allows such objects to be relit.  
Other methods directly predict the physical properties of the surface of a material through learning, using diffusion \cite{vecchio2023matfuse,vecchio2023controlmat} or learning good latent representations from data \cite{zsolnai2018gaussian,henzler2021generative,shimizu2020design,liu2018perception,hu2022inverse,chen2022tango}. 
However, for all of the work mentioned so far, their fundamentally image-based or latent-based representations make the output materials difficult to edit in existing 3D graphics pipelines.
People have also explored combining learning-based approaches with symbolic representations of materials \cite{ritchie2023neurosymbolic,tchapmi2022generating}. 
These works often involve creating differentiable representations of the procedural material graph often used in 3D graphical design pipelines, and backpropagating gradients throughout the graph to produce an image that can match the target\cite{shi2020match,guerrero2022matformer,hu2022node}.
Other works like Infinigen~\cite{raistrick2023infinite} 
use rule-based procedural generation over a library of procedural materials.
However, no prior works in this direction have demonstrated the ability to edit procedural material graphs using \textit{user intention specified by language}~\cite{ritchie2023neurosymbolic}, a task that we are particularly interested in.
Though the aforementioned works excel at material design, they aren't generalizable to other 3D graphics design task settings. BlenderAlchemy, on the other hand, aims to produce a system that can perform various design tasks according to user intents.
This usually requires combining different methods together in a non-trivial way.

\vspace{-1em}\paragraph{\textbf{LLM as general problem solvers.}}
Large language models (LLMs) like GPT-4~\cite{2023GPT4,2023GPT4VisionSC}, Llama~\cite{touvron2023llama}, and Mistral~\cite{jiang2023mistral} have in recent years demonstrated unprecedented results in a variety of problem settings, like robotics \cite{hu2023toward,zhou2023language,firoozi2023foundation,xiao2023robot,singh2023progprompt,liang2023code,ahn2022can}, program synthesis \cite{olausson2023self,shinn2023reflexion,romera2024mathematical}, and graphic design \cite{huang2023aladdin,wen2023anyhome,yang2024holodeck}. 
Other works have shown that by extending such models with an external process like Chain-of-Thought~\cite{wei2022chain}, Tree-of-Thought~\cite{yao2024tree} or memory/skill database~\cite{park2023generative,patil2023gorilla,de2023llmr,wang2023voyager}, or by embedding such systems within environments where it can perceive and act \cite{wang2023voyager,schick2024toolformer,park2023generative,de2023llmr}, a range of new problems that require iterative refinement can be solved. 
Their application to visual problem settings, however, has mostly been limited due to the nonexistent visual perception capabilities of the LLMs \cite{goel2023iterative,wen2023anyhome,yang2024holodeck,yamada2024l3go,de2023llmr,wang2023voyager}. 
While this could be sidestepped by fully condensing the visual state of the environment using text \cite{park2023generative} or some symbolic representation \cite{de2023llmr,wang2023voyager}, doing so for 3D graphic design works poorly.
For instance, the task of editing a Blender material graph to create a desired material requires many trial-and-error cycles and an accurate understanding of the consequences certain design actions can have on the visual output. 
Recent works that apply LLMs to graphical design settings \cite{goel2023iterative,de2023llmr} and ones that more specifically do so within Blender (like BlenderGPT~\cite{blendergpt}, 3D-GPT\cite{sun20233dgpt} and L3GO \cite{yamada2024l3go}) do not use visual information to inform or refine their system's outputs, leading to unsatisfactory results.
BlenderAlchemy borrows ideas from existing LLM literature and tries to address this issue by inputting visual perception into the system.

\vspace{-1em}\paragraph{\textbf{Vision-Language Models.}}
State-of-the-art vision language models, such as LLaVA~\cite{li2024llava}, GPT-4V~\cite{2023GPT4VisionSC}, and Gemini~\cite{team2023gemini} have demonstrated impressive understanding of the visual world and its connections to language and semantics, enabling many computer vision tasks like scene understanding, visual question answering, and object detection to be one API call away \cite{yang2023dawnofllms,yin2023survey,fu2023mme,yin2023woodpecker,fu2023gemini}. 
Works such as \cite{wu2024gpt,baumli2023vision} suggest that such models can also be used as a replacement for human evaluators for a lot of tasks, positioning them as tools for guiding planning and search by acting as flexible reward functions. 
BlenderAlchemy takes the first steps to apply VLMs to solve 3D graphic design tasks, a novel yet challenging application rather unexplored by existing works.

\section{Method}
\label{sec:method}

The goal of our system is to perform edits within the Blender 3D graphic design environment through iteratively refining programs that define a sequence of edits in Blender. This requires us to (1) decompose the input initial Blender input into a combination of programs and a ``base'' Blender state (Section \ref{sec:blender2programs}) and (2) develop a procedure to edit each program to produce the desired visual state within Blender to match a user intention (Sections \ref{sec:iterative_refinement}).


\subsection{Representation of the Blender Visual State} \label{sec:blender2programs}
The state of the initial Blender design environment can be decomposed into an ``base'' Blender state $S_{\text {base}}$ and a set of initial programs $\{p_{0}^{(1)}, p_{0}^{(2)}, ..., p_{0}^{(k)}\}$ that acts on state $S_{\text{base}}$ to produce the initial Blender environment through a dynamics function $F$ that transitions from one state to another based on a set of programmatic actions:

$$ S_{\text{init}} = F\left(\left\{p_{0}^{(i)}\right\}_{i=1...k}, S_{\text{base}}\right)$$

In our problem setting, $F$ is the python code executor within the Blender environment that executes $\{p_{0}^{(i)}\}_{i=1...k}$ in sequence. We set each initial program $p_{0}^{(i)}$ to be a program that concerns a single part of the 3D graphical design workflow -- for instance, $p_0^{(1)}$ is in charge of the material on one mesh within the scene, and $p_0^{(2)}$ is in charge of the lighting setup of the entire scene. The decomposition of $S_{\text{init}}$ into $S_{\text{base}}$ and $ p_{0}^1...p_{0}^k$ can be done using techniques like the ``node transpiler'' from Infinigen~\cite{raistrick2023infinite}, which converts entities within the Blender instance into lines of Python code that can recreate a node graph, like a material shader graph. We develop a suite of tools to do this in our own problem setting. 

Though it's possible for \textit{all} edits to be encompassed in a \textit{single} program instead of $k$ programs, this is limiting in practice -- either because the VLM's output length isn't large enough for the code necessary or because the VLM has a low success rate, due to the program search space exploding in size. Although it is possible that future VLMs will substantially mitigate this problem, splitting the program into $k$ task-specific programs may still be desirable, given the possibility of querying $k$ task-specific fine-tuned/expert VLMs in parallel.

\subsection{Iterative Refinement of Individual Visual Programs} \label{sec:iterative_refinement}

Suppose that to complete a task like material editing for a single object, it suffices to decompose the initial state into $S_{\text{base}}$ and a single script, $p_0$ -- that is, $S_{\text{init}} = F(\{p_0\}, S_{\text{base}})$. Then our goal is to discover some edited version of $p_0$, called $p_1$, such that $F(\{p_1\}, S_{\text{base}})$ produces a visual state better aligned with some user intention $I$. Our system assumes that the user intention $I$ is provided in the form of language and/or image references, leveraging the visual understanding of the latest VLM models to understand user intention.

\begin{algorithm}
\caption{Iterative Refinement of Visual Programs}
\label{alg:iterative_refinement}
\begin{algorithmic}[1]
\Procedure{Tournament}{State candidates $\{S_1, S_2, ...S_k\}$, Visual state evaluator $V$, User intention $I$}
    \If{$k>2$}
        \State $w_1 \gets$ \textsc{Tournament($\{S_1, S_2, ...S_{k/2}\}$, $V$, $I$)}, 
        \State $w_2 \gets$ \textsc{Tournament($\{S_{k/2}, S_{k+1}, ...S_{k}\}$, $V$, $I$)}
    \Else
        \State $w_1 \gets S_1, w_2 \gets S_2$
    \EndIf
    \State \Return $V(w_1, w_2, I)$
\EndProcedure

\Procedure{Refine}{Depth $d$, Breadth $b$, Intention $I$,
                   Edit Generator $G$, State Evaluator $V$,
                   Base state $S_{\text{base}}$,
                   Initial program $p_0$,
                   Dynamics Function $F$}
    \State $S_0 \gets F(p_0, S_{\text{base}})$, $S_{\text{best}} \gets S_0$, $p_{\text{best}} \gets p_0$
    \For{$i \gets 1$ to $d$}
        \State $\mathcal P_i = \mathcal N(p_{\text{best}})$ if $i$ is odd else $\mathcal P$ \Comment{``Tweak'' or  ``Leap'' edits} \label{alg:line:tuneleap}
        \State $p_{i}^{1}, p_i^2 ... p_i^b \gets G(p_{\text{best}}, S_{\text{best}}, I, b, \mathcal P_i)$ \Comment {Generate $b$ options}
        \State $S_{i}^1 \gets F(p_{i}^1, S_0), ..., S_{i}^b \gets F(p_{i}^b, S_0)$ \Comment {Observe the visual states}
        \State $S_{i}^* \gets$ \textsc{Tournament(}$\{S_{i}^1, S_{i}^2 ...  S_{i}^b \} \cup \{S_{\text{best}}\}$, $V$, $I$\textsc{)} \Comment {Choose the best} \label{alg:line:rebasing_candidate_tournament}
        \State $S_{\text{best}} \gets S_{i}^*$, $p_{\text{best}} \gets p_{i}^*$ \Comment{Best visual state and programs so far}
    \EndFor
    \State \Return $S_{\text{best}}, p_{\text{best}}$
\EndProcedure
\end{algorithmic}
\end{algorithm}

To discover a good edit to $p_0$, we introduce the procedure outlined in Algorithm \ref{alg:iterative_refinement}, an iterative refinement loop that repeatedly uses a visual state evaluator $V$ to select among the hypotheses from an edit generator $G$. A single ``agent'' for a certain task like procedural material design can be fully described by $(G, V)$.

Inspired by works like \cite{wu2024gpt}, we propose a visual state evaluator $V(S_1, S_2, I)$, which is tasked with returning whichever of the two visual states ($S_1$ or $S_2$) better matches the user intention $I$. This evaluator is applied recursively to choose the most suitable visual state candidate among $b$ visual state candidates by making $\mathcal O(\log(b))$ queries, as done in \textsc{Tournament} in Algorithm \ref{alg:iterative_refinement}.

Though it seems straightforward to ask the same VLM to edit the code in a single pass, this leads to many failure cases (Section \ref{sec:ablations}). 
Due to the VLM's lack of baked-in understanding of the visual consequences of various programs within Blender, a multi-hypothesis and multi-step approach is more appropriate.
Extending Tree-of-Thoughts \cite{yao2024tree} to the visual domain, $G(p, S, I, b, \mathcal P)$ is a module tasked with generating $b$ different variations of program $p$, conditioned on the current visual state $S$ and user intention $I$, constrained such that the output programs fall within some family of programs $\mathcal P$, which can be used to instill useful priors to the edit generator.

Below we describe some additional system design decisions that ensured better alignment of the resultant edited program to the user intention, either by improving the stability of the procedure or by supplementing the visual understanding of VLMs. The effect of each is investigated in Section \ref{sec:ablations}.

\paragraph{\textbf{Hypothesis Reversion.}}
To improve the stability of the edit discovery process, we add the visual state of the program being edited at every timestep ($S_{\text{best}}$ for $p_{\text{best}}$) as an additional candidate to the selection process, providing the option for the process to revert to an earlier version if the search at a single iteration was unsuccessful. Line \ref{alg:line:rebasing_candidate_tournament} in Algorithm \ref{alg:iterative_refinement} shows this. 

\paragraph{\textbf{Tweak and Leap Edits.}}\label{sec:tuneleap}
An important characteristic of visual programs is that continuous values hard-coded within the program can modify the output just as much as structural changes. This is in contrast to 
non-visual program synthesis tasks based on unit-tests of I/O specs, like \cite{chen2021codex,austin2021program}, where foundation models are mostly tasked to produce the right \textit{structure} of the program with minimal hard-coded values.
Given a a single visual program $p \in \mathcal P$, the space of visual outputs achievable through \textit{only} tweaking the numerical values to function parameters and variable assignments can cover a wide range, depending on the fields available in the program. 
In Algorithm \ref{alg:iterative_refinement}, we refer this space as the ``neighborhood'' of $p$ or ``tweak'' edits, $\mathcal N(p)$. Though this can result in a very small change to the program, this can lead to a large visual difference in the final output, and in a few edits can change the ``wrong'' program into one more aligned with the user intention. 
On the other hand, more drastic changes (or ``leap'' edits) may be needed to accomplish a task. Consider for example the task of changing a perfectly smooth material to have a noised level of roughness scattered sparsely across the material surface. This may require the programmatic addition of the relevant nodes (e.g.\ color ramps or noise texture nodes), and thus the edited program falls  outside of $\mathcal N(p)$.

Empirically, we find that the optimal edits are often a mix of tweak and leap edits. As such, our procedure cycles between restricting the edits of $G$ to two different sets: the neighborhood of $p_{\text{best}}$, and the whole program space $\mathcal P$ (Line \ref{alg:line:tuneleap} in Algorithm \ref{alg:iterative_refinement}). In practice, such restrictions are softly enforced through in-context prompting of VLMs, and though their inputs encourage them to abide by these constraints, the model can still produce more drastic ``tweak'' edits or conservative ``leap'' edits as needed.
 
\paragraph{\textbf{Visual Imagination.}}\label{sec:visual_imagination}
In the case when the user intention is communicated purely textually, their intention may be difficult for the VLM to turn into successful edits.
Prior works like \cite{huang2023aladdin,wen2023anyhome,yang2024holodeck} have made similar observations of abstract language for 3D scenes. Consider, for example, the prompt ``make me a material that resembles a \textit{celestial nebula}''. To do this, the VLM must know what a celestial nebula looks like, \textit{and} how it should change the parameters of the material shader nodes of, say, a wooden material. We find that in such cases, it's hard for the VLM to directly go from abstract descriptions to low-level program edits that affect low-level properties of the Blender visual state. 

Instead, we propose supplementing the text-to-program understanding of VLM's with the text-to-image understanding in state-of-the-art image generation models. Intermediary visual artifacts (images generated using the user intention) are created and used to guide the refinement process towards a more plausible program edit to match the desired outcome, as shown in Figure \ref{fig:teaser}. The generated images act as image references \textit{in addition to the textual intention provided by the user}. This constitutes a simple visual chain of thought~\cite{wei2022chain} for visual program editing, which not only creates a reference image for $G$ and $V$ to guide their low-level visual comparisons (e.g. color schemes, material roughness... \etc), but also provides a user-interpretable intermediary step to confirm the desired goals behind an otherwise vague user intention.


\section{Experiments}
\label{sec:experiments}
We demonstrate BlenderAlchemy on editing procedural materials, geometry and lighting setups within Blender, three of the most tedious parts of 3D design. 

\subsection{Procedural Material Editing}

Procedural material editing has characteristics that make it difficult for the same reason as a lot of other visual program settings: (1) small edit distances of programs may result in very large visual differences, contributing to potential instabilities in the edit discovery process, (2) it naturally requires trial-and-error when human users are editing them as the desired magnitude of edits depends on the visual output and (3) language descriptions of edit intent typically do not contain low-level information for the editor to know immediately which part of the program to change.

We demonstrate this on two different kinds of editing tasks: (1) turning the same initial material (a synthetic wooden material) into other materials described by a list of \textbf{language descriptions}  and (2) editing many different initial materials to resemble the same target material described by an \textbf{image input}. 
For the starting materials, we use the synthetic materials from Infinigen~\cite{raistrick2023infinite}. 

\subsubsection{Text-based material-editing}

An attractive application of our system is in the modification of preexisting procedural materials using natural language descriptions that communicate user intent, a desireable but thus far undemonstrated capability of neurosymbolic methods~\cite{ritchie2023neurosymbolic}. We demonstrate our system's capabilities by asking it to edit a wooden material from Infinigen~\cite{raistrick2023infinite} into many other materials according to diverse language descriptions of target materials that are, importantly, \textit{not wood}. In reality, this is a very challenging task, since this may require a wide range in the size of edits even if the language describing desired target material may be very similar. 

\begin{figure}[tb]
    \centering
    \includegraphics[width=\textwidth]{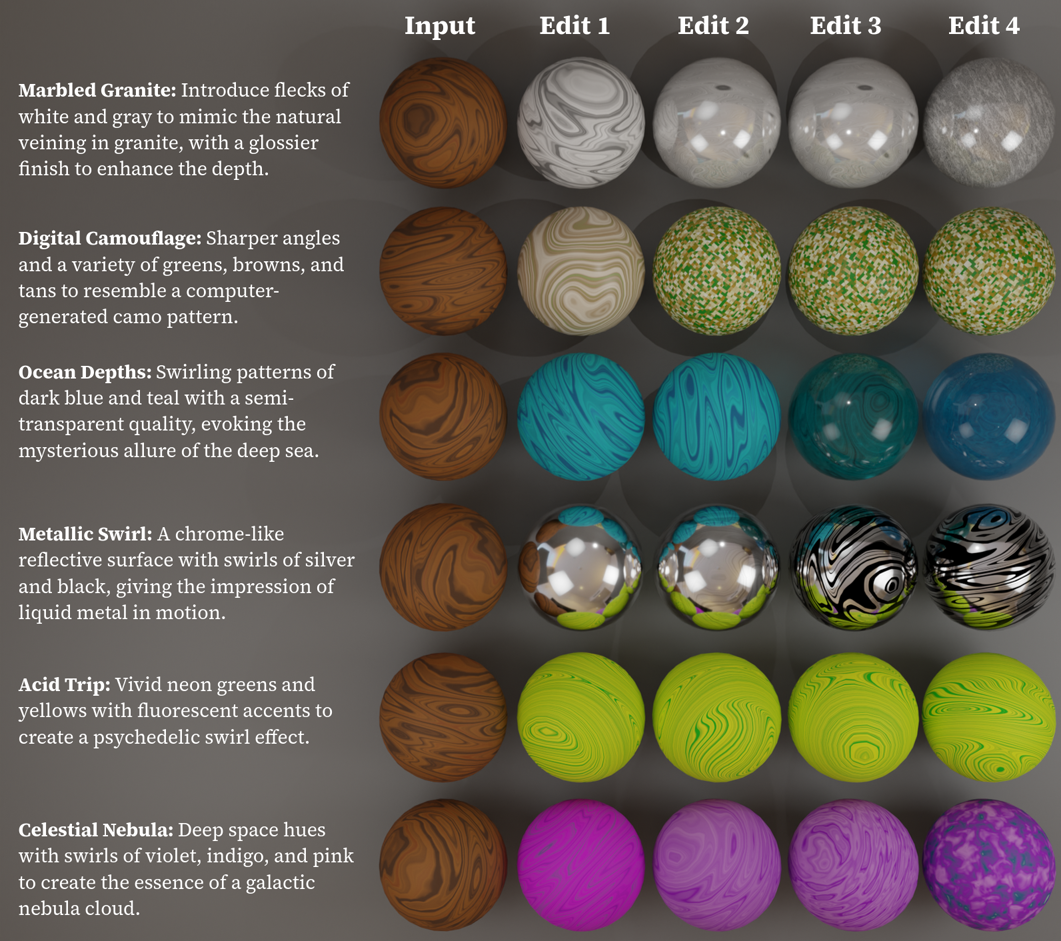}
    \vspace{-2em}
    \caption{\textbf{Text-based Material Editing Results.} The step-by-step edits of a 4x8 version of BlenderAlchemy to the same wooden material, given the text description on the left as the input user intention.}
    \label{fig:wood2others}
    \vspace{-2em}
\end{figure}

\begin{figure}[tb]
    \centering
    \includegraphics[width=\textwidth]{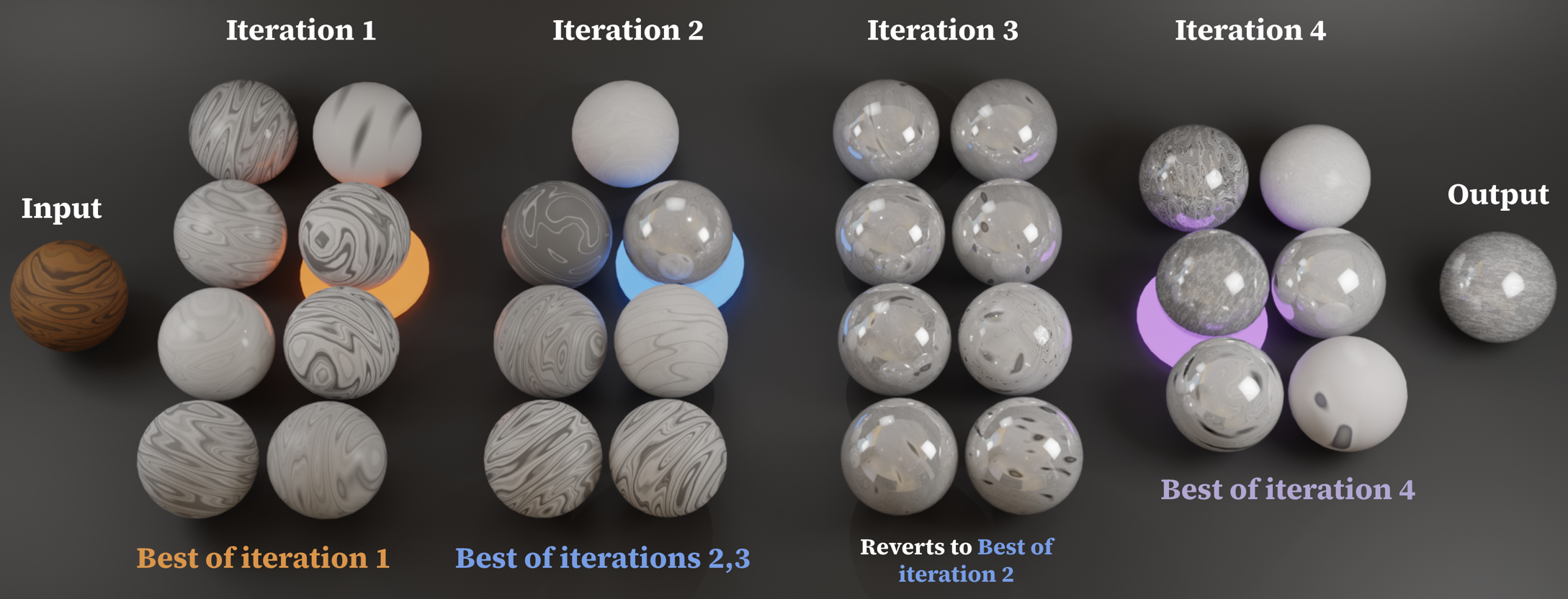}
    \vspace{-2em}
    \caption{\textbf{The edit discovery process of turning a wooden material into ``marbled granite''.} Each column shows the hypotheses generated by $G$, with the most promising candidates chosen by $V$ indicated by the highlights. Note that iteration 3 proved to be unfruitful according to $V$, and the method reverts to the best candidate from iteration 2, before moving onto iteration 4. }
    \label{fig:wood2marble_process}
\end{figure}

Figure \ref{fig:wood2others} shows examples of edits of the same starter wood material using different language descriptions, and Figure \ref{fig:wood2marble_process} demonstrates the intermediary steps of the problem solving process for a single instance of the problem. Our system is composed of an edit generator that generates 8 hypotheses per iteration, for 4 iterations ($d=4, b=8$), cycling between tweak and leap edits. It uses GPT-4V for edit generation and state evaluation, and DallE-3 for visual imagination.

We compare against BlenderGPT, the most recent open-sourced Blender AI agent that use GPT-4 to execute actions within the Blender environment through the Python API. \footnote{At the time of publishing, works like 3DGPT\cite{sun20233dgpt}, L3GO\cite{yamada2024l3go} have not yet open-sourced their code.}
We provide the same target material text prompt to BlenderGPT, as well as the starter code for the initial wood material for reference.
We compare the CLIP similarity of their output material to the input text description against our system. BlenderGPT reasons only about how to edit the program using the input text description, doing so in a single pass without state evaluation or multi-hypothesis edit generation. To match the number of edit generator queries we make, we run their method a maximum of 32 times, using the first successful example as its final output. Everything else remains the same, including the starter material program, text description, base Blender state, and lighting setup.

\begin{figure}[tb]
    \centering
    \includegraphics[width=\textwidth]{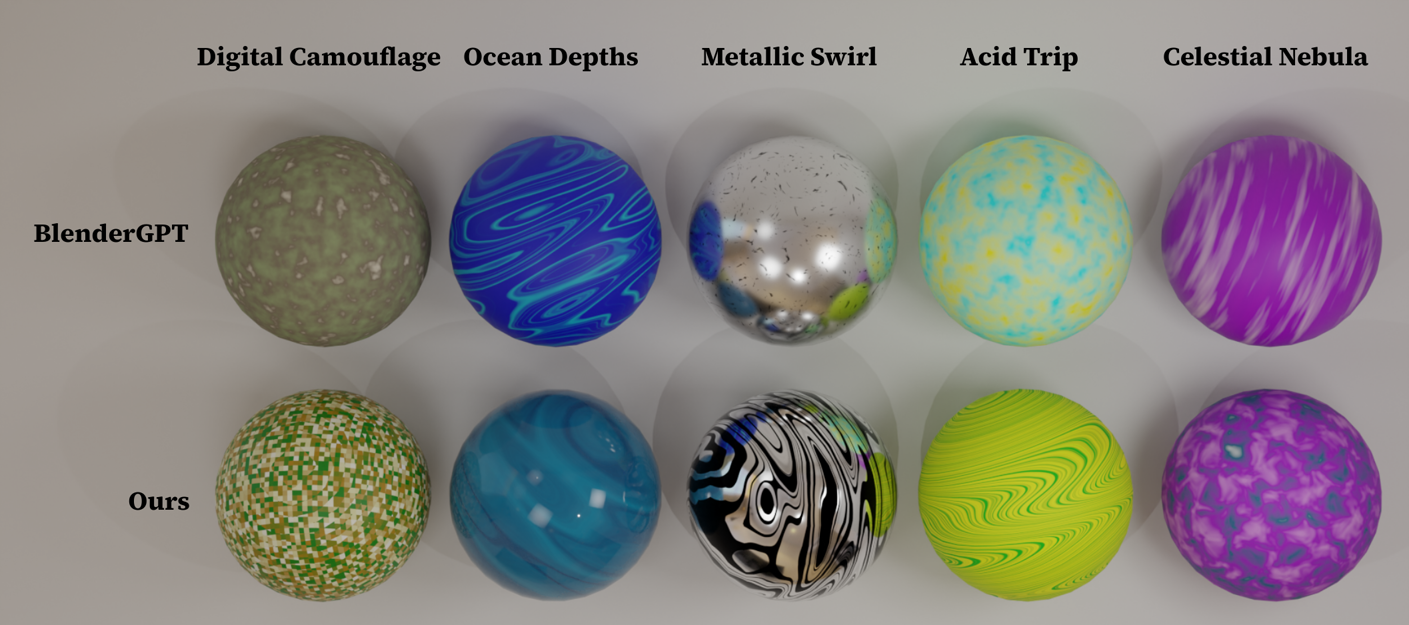}
    \vspace{-2em}
    \caption{\textbf{Comparisons between our method and BlenderGPT for the text-based material editing task setting.} Note how our materials align better with the original language prompts (See Figure \ref{fig:wood2others} for the original prompts). The input material being edited is the same wooden material.}
    \label{fig:wood2others_vsbaseline}
    
\end{figure}

We find that qualitatively, BlenderGPT produces much shallower and more simplistic edits of the input material, resulting in low-quality output materials and poor alignment with the user intention. Examples can be seen in Figure~\ref{fig:wood2others_vsbaseline}. For instance, observe that for the ``digital camouflage'' example, BlenderAlchemy is able to produce the ``sharper angles'' that the original description requests (See Figure~\ref{fig:wood2others}) whereas BlenderGPT produces the right colors but fails to create the sharp, digital look. For ``metallic swirl'' example, our system's visual state selection process would weed out insufficiently swirly examples such as the one given by BlenderGPT, enabling our method to produce a material closer to the prompt.

Table~\ref{tab:baseline_comparisons} demonstrates the comparison in terms of the average ViT-B/32 and ViT-L/14 CLIP similarity scores~\cite{radford2021learning} with respect to the language description. Our system's ability to iteratively refine the edits based on multiple guesses at each step gives it the ability to make more substantive edits over the course of the process. Moreover, visual grounding provided both by the visual state evaluator as well as the output of the visual imagination stage guides the program search procedure to better align with the description. 

We additionally conducted a user preference study to compare BlenderAlchemy's performance with two material 
generation baselines that use diffusion: TEXTure~\cite{richardson2023texture} and Paint3D~\cite{zeng2023paint3d}.
We collect 592 
Mechanical Turk comparisons between BlenderAlchemy and the baselines
from 24 Turkers on materials created using 32 different text prompts.
Users must choose the material that best matches the text-prompt.
The results show that BlenderAlchemy is preferred to Paint3D \textit{73\%} of the time.
Users picked BlenderAlchemy over TEXTure \textit{56\%} of the time.
We found that in \textit{71\%} of comparisons, human users also prefer 
BlenderAlchemy outputs selected by our visual evaluator.
This suggests that our visual evaluator makes decisions that align with users' preferences.
\vspace{-1em}

\begin{table}[tb]
    \caption{\textbf{CLIP scores of BlenderAlchemy vs. BlenderGPT} for the text-based material editing task. We find that a version of our system that has \textit{no visual components} (\textbf{-Vision}) still outperforms BlenderGPT. By adding vision to the state evaluator alone (\textbf{-Vision G}) and not the edit generator, a further improvement is observed.}
    \label{tab:baseline_comparisons}
    \centering
    \begingroup
    \setlength{\tabcolsep}{5pt} 
    \renewcommand{\arraystretch}{1.} 
    \begin{tabular}{@{\hskip 0.5em}ccccc@{\hskip 0.5em}}
    \toprule
    Metric & BlenderGPT & -Vision & -Vision G & Ours \\ \midrule
    ViT-B/32 ($\uparrow$) & 25.2       & 25.7          & 27.8            & \textbf{28.2} \\
    ViT-L/14 ($\uparrow$) & 21.1       & 21.8          & 23.4            & \textbf{24.0}  \\
    \bottomrule
    \end{tabular}
    \endgroup
    \vspace{-1em}
\end{table}

\subsubsection{Image-based material-editing}

Given an image of a desired material, the task is to convert the code of the starter material into a material that contains many of the visual attributes of the input image, akin to doing a kind of style transfer for procedural materials.

\begin{figure}[tb]
    \centering
    \includegraphics[width=\textwidth]{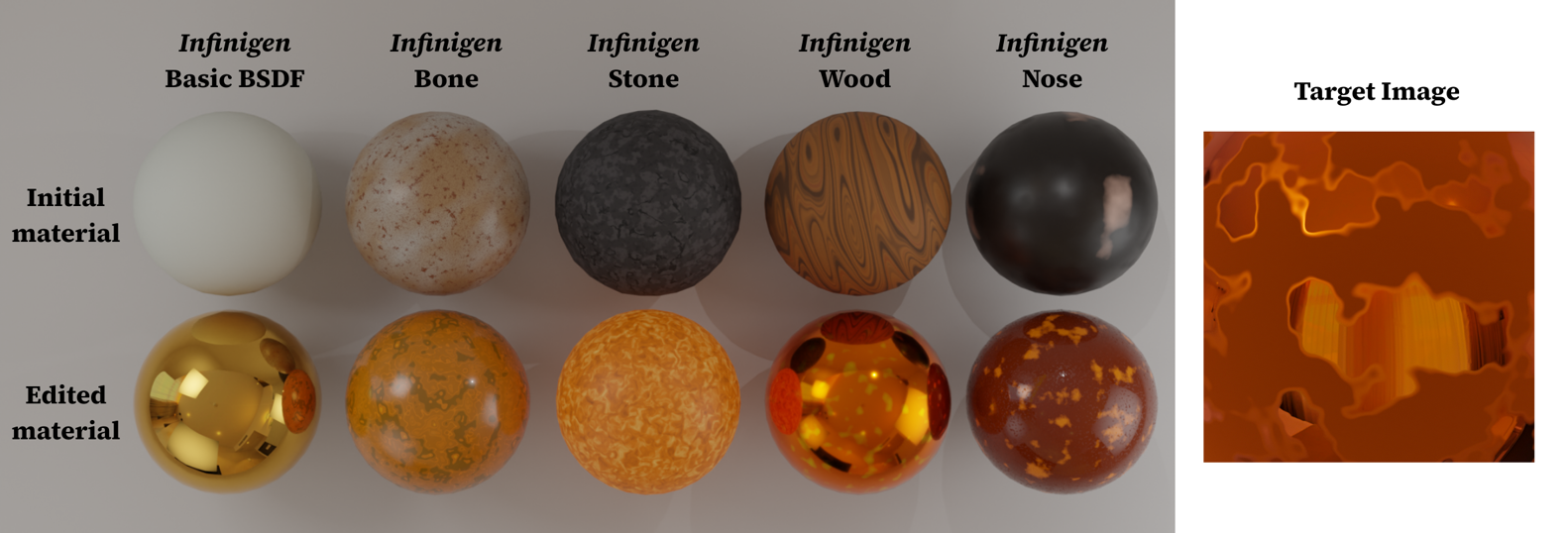}
    \vspace{-2em}
    \caption{\textbf {Material editing based on image inputs.} Our edit intention is described by the target image shown on the right. 5 different Infinigen~\cite{raistrick2023infinite} initial materials are shown here, and the final edit. Note how in each case, certain attributes of the target material (metallicness, color, texture) are transferred.}
    \label{fig:infinigen2metal}
    \vspace{-1.5em}
\end{figure}

At each step, the edit generator is first asked to textually enumerate a list of obvious visual differences between the current material and the target, then asked to locate lines within the code that may be responsible for these visual differences (\eg ``the target material looks more rough'' $\rightarrow$ ``line 23 sets a roughness value, which we should try to increase'') before finally suggested an edited version of the program. As such, low-level visual differences (\eg color discrepancies) are semantically compressed first into language descriptions (\eg ``the target is more red''), before being fed into the editing process, resulting in the behavior that our system produces variations of the input material that resembles the target image along many attributes, even if its outputs don't perfectly match the target image (See Figure \ref{fig:infinigen2metal}).
Our system is the same as for text-based material editing, but without the need for visual imagination.
\vspace{-1em}

\section{Ablation Experiments on Material Editing} \label{sec:ablations}


\subsubsection{Value of visually-grounded edit generators and state evaluators}
How important is it that the edit generator $G$ and state evaluator $V$ have access to vision? We ablate (1) visual perception of $G$ by prompting it to \textit{only} propose edits based on the text of the initial program $p_0$, without \textit{any visual information} (neither the initial rendering $S_0$ or, if applicable, images of intentions) and (2) additionally the visual perception of $V$, by letting it decide which of two programs is better purely on the basis of the program code and the textual description of the user intention $I$. Table \ref{tab:imag_rebase_tl_ablation} shows the consequences of such ablations to the alignment to the user intention. 
\vspace{-1em}
\subsubsection{Value of the multi-hypothesis, multi-step approach}
We vary the dimensions of our system ($d$ and $b$ within Algorithm 1 in the main paper 
) to demonstrate the importance of having a well-balanced dimensions. We keep the total number of calls to the edit generator constant ($d \times b$) at 32 requests. Table \ref{tab:dimensions_ablations} shows the result of this on the image-based material editing task. This suggests that there's a sweet-spot in the system dimensions ($8 \times 4$ or $4 \times 8$) where the system is neither too wide (large $b$, in which case the system tends to overly explore the program space, and produce insubstantial edits) or too tall (large $d$, and in which case the system has high risk of overly exploiting suboptimal edit candidates). By the same argument, even if BlenderGPT~\cite{blendergpt} was equipped with visual perception and a state evaluator to choose among 32 candidates, it would still suffer from the same issues as the $1\times32$ version of our method.
\begin{table}[tb]
    \caption{\textbf{Different dimensions of BlenderAlchemy and their metrics on the image-based material editing task}. ``$1 \times 32$'' indicates a setup that uses $d=1$ (number of iterations) and $b=32$ (number of hypotheses per iteration) in Algorithm 1 in the main paper
    . This shows the  clear advantage of using a more balanced choice of $d$ and $b$ over sequential iterative refinement method ($32 \times 1$) or querying the language model multiple times without refinement ($1\times32$).}
    \label{tab:dimensions_ablations}
    \centering
    \begingroup
    \setlength{\tabcolsep}{5pt} 
    \renewcommand{\arraystretch}{1.} 
    \begin{tabular}{@{}lcccccc@{}}
    \toprule
                & $1\times 32$ & $2\times 16$ & $4\times 8$    & $8 \times 4$  & $16\times 2$ & $32\times1$ \\ \midrule
    ViT-B/32 ($\uparrow$)    & 81.7         & 82.9         & 81.7           & \textbf{84.1} & 83.6         & 81.6        \\
    Photometric ($\downarrow$)  & 0.066        & 0.066        & \textbf{0.049} & 0.050         & 0.056        & 0.087       \\
    LPIPS ($\downarrow$)  & 0.64         & 0.54         & 0.52           & \textbf{0.50} & 0.52         & 0.59        \\ \bottomrule
    \end{tabular}
    \endgroup
\end{table}
\begin{table}[tb]
    \caption{\textbf{Ablating system design decisions.} For the text-based material editing task, we compare against variants in which we remove (1) visual perception from $G$ \textit{and} $V$ (\textbf{-Vision}), (2) visual perception from $G$ and \textit{not} from $V$ (\textbf{-Vision G}), (3) visual imagination (\textbf{-Imagin.}), (4) reversion capabilities (\textbf{-Revert}), (5) the option of leap edits (\textbf{-Leap}) or (6) tweak edits (\textbf{-Tweak}). We use a $4 \times 8$ version of BlenderAlchemy. Edits 1 to 4 correspond to the output at each refinement step of the BlenderAlchemy process. We show the ViT-B/32 CLIP scores here.}
    \label{tab:imag_rebase_tl_ablation}
    \centering
    \begingroup
    \setlength{\tabcolsep}{5pt} 
    \renewcommand{\arraystretch}{1.} 
    \begin{tabular}{@{\hskip 1em}l@{\hskip 1em}ccccccc@{\hskip 1em}}
        \toprule 
        & -Vision & -Vision G & -Imagin. & -Revert & -Leap & -Tweak & Ours \\
        \midrule
        Edit 1 & 27.4       & 27.6             & 26.8          & 27.1     & 27.1   & 27.2    & \textbf{27.8} \\
        Edit 2 & 26.1       & 27.6             & 27.1          & 26.4     & 27.6   & 27      &  \textbf{27.9} \\
        Edit 3 & 26.5       & 27.6             & 26.8          & 26.6     & 27.7   & 26.9    & \textbf{28.4} \\
        Edit 4 & 25.7       & 27.8             & 26.9          & 25.8     & 27.8   & 26.6    & \textbf{28.2} \\ 
        \bottomrule
    \end{tabular}
    \endgroup
\end{table}
\vspace{-1em}
\subsubsection{Hypothesis Reversion}

The intended effect of hypothesis reversion is to ensure the stability of the procedure, especially when (1) the tree's depth is sufficient for the edit search to go astray and (2) when leap edits can cause large and potentially disruptive edits in a single iteration of our procedure. As seen in Table \ref{tab:imag_rebase_tl_ablation}, removing the ability to revert hypotheses causes divergence of the alignment to the text description over several edits, and larger drops in average CLIP-similarity corresponds to when leap edits happen (Edit 2 and Edit 4). 

\vspace{-1em}
\subsubsection{Tweak/Leap Edits}
When we ablate all ``tweak'' edits (``-Tweak'' in Table \ref{tab:imag_rebase_tl_ablation}) by making all edits ``leap'' edits, we observe a strong divergence from the user intention. Conversely, ablating all ``leap'' edits (``-Leap'' in Table \ref{tab:imag_rebase_tl_ablation}) leads to slow but steady increase in alignment with the user intention, but too conservative to match the ``tweak+leap'' variant (``Ours'' in Table \ref{tab:imag_rebase_tl_ablation}). 

\vspace{-1em}
\subsubsection{Visual Imagination}
Visual imagination is an additional image-generation step before launching the procedure in Algorithm 1 in the main paper
, with the intended effect of guiding the edit generator and state evaluator with text-to-image understanding of state of the art image diffusion models. Without it, user intentions communicated using abstract language descriptions lead to poorer edits due to having limited information to properly guide the low-level visual comparisons (e.g. color, textures, ...\etc) by the state evaluator and edit generator (See Table \ref{tab:imag_rebase_tl_ablation}).


\begin{figure}[t]
  \centering
   \includegraphics[width=\linewidth]{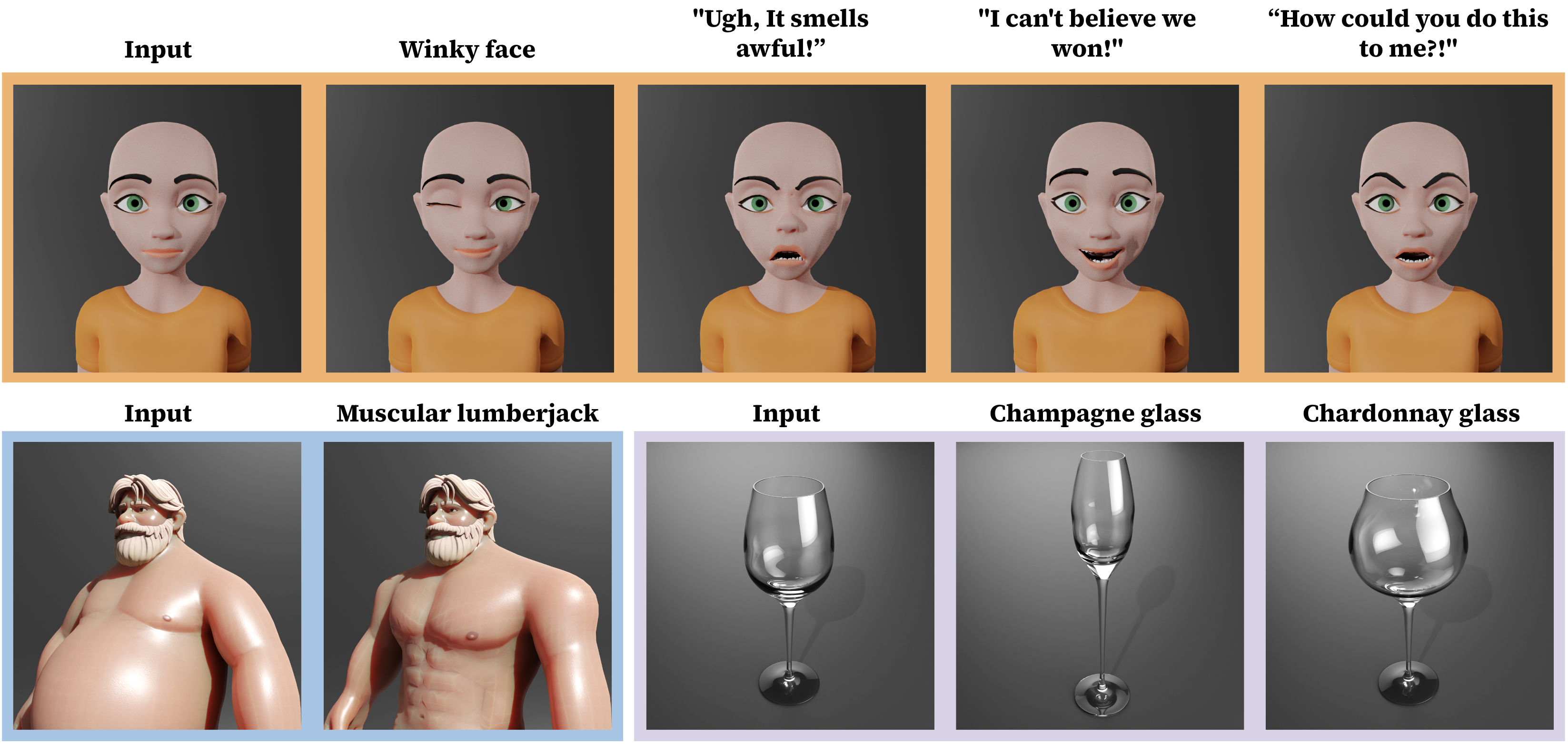}
   \vspace{-1.5em}
   \caption{BlenderAlchemy editing geometry using blend shapes. Edits are made to match a description or a script line. 
   Input shapes from BlenderKit.}
   \label{fig:shapekeys}
   
\end{figure}
\begin{figure}[t]
  \centering
   \includegraphics[width=\linewidth]{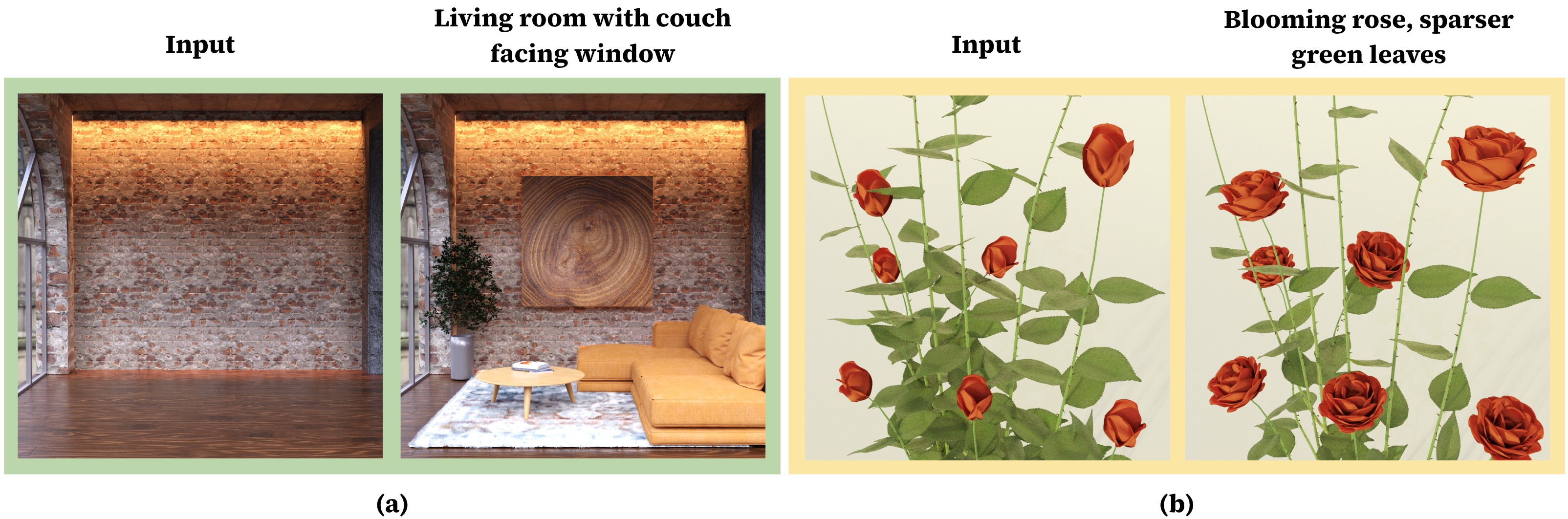}
   \vspace{-1.5em}
   \caption{BlenderAlchemy editing \textbf{(a)} the placement of assets within a living room scene and \textbf{(b)} the procedural geometry nodes of roses, to different text prompts.}
   \label{fig:geonodes_placement}

\end{figure}
\begin{figure}[tb]
    \centering
    \includegraphics[width=\linewidth]{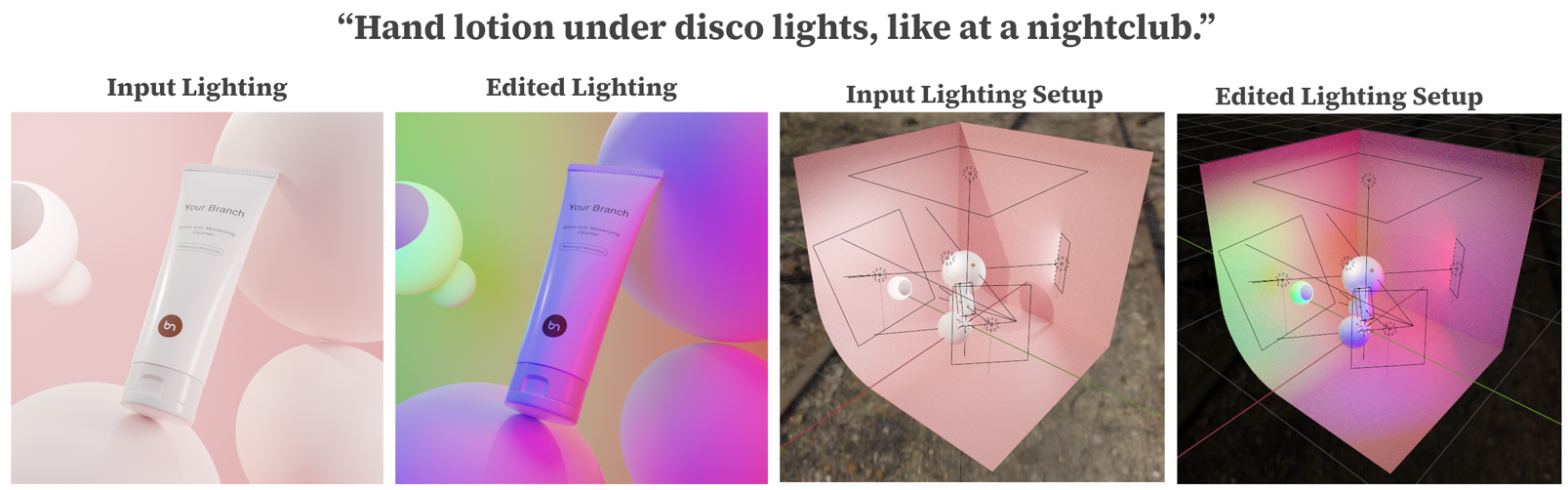}
    \vspace{-2em}
    \caption{\textbf{Optimizing the lighting} of the scene setup to the text-based user intention. We base the initial Blender state input based on a product visualization downloaded from BlenderKit.}
    \label{fig:disco_lotion}
    \vspace{-1em}
\end{figure}

\begin{figure}[tb]
    \centering
    \includegraphics[width=\textwidth]{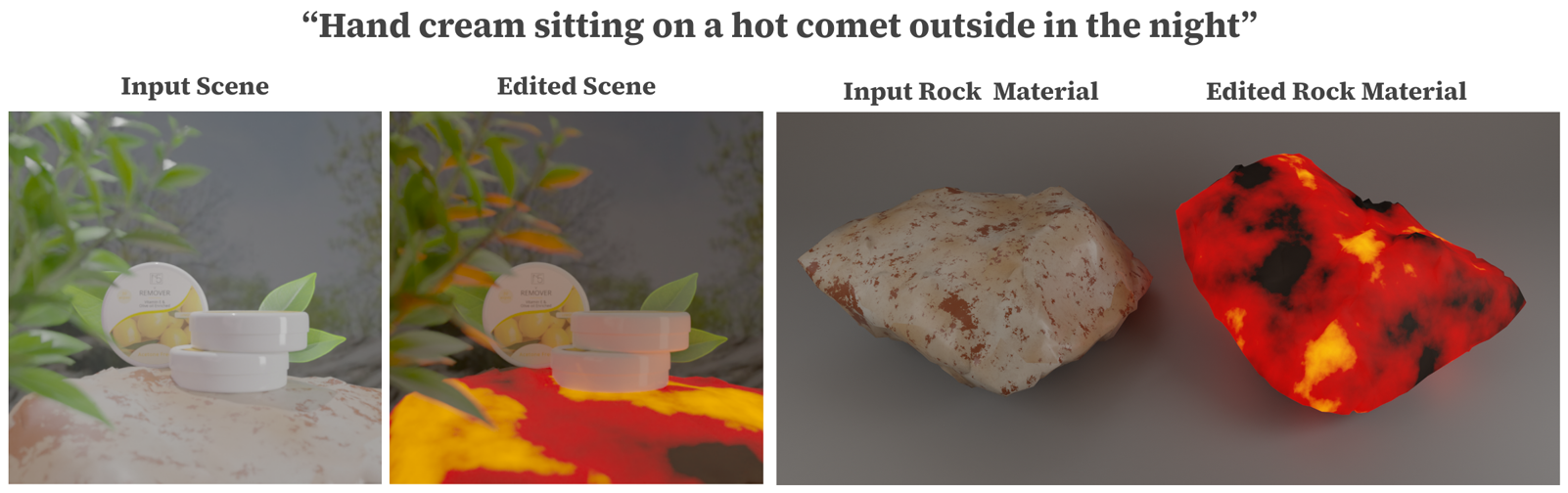}
    \vspace{-2em}
    \caption{\textbf{Optimizing the lighting and material iteratively} within a product visualization scene, to satisfy the text-based user intention. Note the dimming of the environment lighting for the nighttime lighting, and the glowing-hot material the editing procedure has produced. We base the initial Blender state input based on a product visualization downloaded from BlenderKit.}
    \label{fig:comet_cream}
    \vspace{-1.5em}
\end{figure}

\subsection{Geometry Editing}


BlenderAlchemy can also be used to manipulate geometry. We show qualitative examples of our system controlling geometry by programmatically 1) interpolating between preset blend shapes, 2) editing of geometry node graphs, and 3) the precise placement of objects within scenes. The blend weights, geometry node graphs and object placements can all be represented programmatically.

As shown in Figure \ref{fig:shapekeys}, given an input shape and pre-defined blend shapes, BlenderAlchemy can modify the shape by finding the appropriate blend weights between different shapes to match language, which is non-trivial for human users given the number of blend shapes 
to interpolate for complex characters and shapes. For instance, the facial expression editing example requires coordinated interpolation between 19 different blend shapes, each which controls a specific kind of motion in one part of the face (e.g. an left eye-brow raise). To aid it in this task, our task description to BlenderAlchemy includes language labels for each blend shape.

In Figure \ref{fig:geonodes_placement}(a), 
BlenderAlchemy is presented with a scene layout editing task, where assets are initially dropped out of camera view and must be iteratively moved to fit the scene of a living room and the requirement that the couch is ``facing the window'', through calling a set of object placement helper functions. Figure \ref{fig:geonodes_placement}(b) shows examples of BlenderAlchemy editing the procedural geometry of a set of roses to match a nuanced instruction that influences the number and angle of pedals, as well as the density of leaves. The full geometry node graph that 
BlenderAlchemy edits contains 60 geometry nodes.
\vspace{-0.5em}

\subsection{Lighting Setup Editing}
We show that BlenderAlchemy can be used to adjust the lighting of scenes according to language instructions as well. Figure \ref{fig:disco_lotion} shows this being done for an input product visualization designed by a Blender artist.

As mentioned in Section \ref{sec:blender2programs}, we can consider iteratively optimizing two separate programs, one controlling the lighting of the whole scene and another controlling the material of an object within the scene. That is, $S_{\text{init}} = F(\{p_0^{(L)}, p_0^{(M)}\}, S_{\text{base}})$ for initial lighting program $p_0^{(L)}$ and material program $p_0^{(M)}$. Figure \ref{fig:comet_cream} shows an example of this, where two separate pairs of edit generators and state evaluators, $(G_M, V_M)$ and $(G_L, V_L)$, are used to achieve an edit to the scene that aligns with the user intention.
Algorithm \ref{alg:multiskill_refinement} 
outlines how that can be done.
\begin{algorithm}
\caption{Optimization of many programs using Algorithm \ref{alg:iterative_refinement}}
\label{alg:multiskill_refinement}
\begin{algorithmic}[1]
\Procedure{MultiskillRefine}{Iteration number $N$, Agent collection $\{( G_1, V_1), (G_2, V_2) ... (G_k, V_k)\}$, Base state $S_{\text{base}}$ and Initial programs $\{p_0^{(1)}, p_0^{(2)}, ...p_0^{(k)}\}$, User intention $I$, Dynamics function $F$}

\State $p_{best}^{(1)} \gets p_0^{(1)}$, ..., $p_{best}^{(k)} \gets p_0^{(k)}$  \Comment{Initialize best program edits}

\For{$i \gets 1$ to $N$}:
    \For{$j \gets 1$ to $k$}:
        \State $F_j \gets \lambda x, S_{\text{base}} : F(\{p_{\text{best}}^{(a)}\}_{a \neq j} \cup \{x\}, S_{base})$ \Comment{partial function}
        \State $S_{\text{best}}^j, p_{\text{best}}^j \gets$ \textsc{Refine(}$d, b, I, G_{j}, V_{j}, S_{\text{base}}, p_{\text{best}}^{(j)}, F_{j}$\textsc{)}
    \EndFor
\EndFor
\State \Return $\{p_{\text{best}}^{(1)}, ...  p_{\text{best}}^{(k)}\}$
\EndProcedure
\end{algorithmic}
\end{algorithm}

\vspace{-1.0em}

\section{Conclusion \& Discussion}
\label{sec:conclusion}
\vspace{-0.5em}
In this paper, we introduce BlenderAlchemy, a system that performs edits within the Blender 3D design environment by leveraging vision-language models to iteratively refining a program to be more aligned with the user intention, by using visual information to both explore and prune possibilities within the program space. We equip our system with visual imagination by providing it access to text-to-image models, a tool it uses to guide itself towards program edits that better align with user intentions. We've demonstrated BlenderAlchemy on editing materials, geometry and lighting, and hope that future works will extend this to other workflows as well.

\vspace{-0.5em}

\section*{Acknowledgements}
We acknowledge the support of ARL grant W911NF-21-2-0104 and a Vannevar Bush Faculty Fellowship.
We'd additionally like to thank Maneesh Agrawala for general discussions, and Purvi Goel, Mika Uy, Vishnu Sarukkai, Fan-yun Sun and Sharon Lee for feedback on paper revisions.

\bibliographystyle{splncs04}
\bibliography{main}

\appendix
\pagebreak
\clearpage

Section \ref{sec:ablate_qualitative} studies the qualitative outcomes of ablation demonstrated in the main paper. We outline the prompts we use for the state evaluator and edit generator in Section \ref{sec:prompts}. In Section \ref{sec:code_analysis}, we provide insights on the code edits that BlenderAlchemy produces, and the resultant changes to the material node graph. Finally, we showcase some renderings of scenes that feature BlenderAlchemy materials in Section \ref{sec:applied_materials}.
Lastly, we discuss the societal impact and limitations of our work in Sections \ref{sec:impact} and \ref{sec:limitations}. 



\section{Explaining System Design Decisions Through Qualitative Examples} \label{sec:ablate_qualitative}

\begin{figure}[tb]
    \centering
    \includegraphics[width=\textwidth]{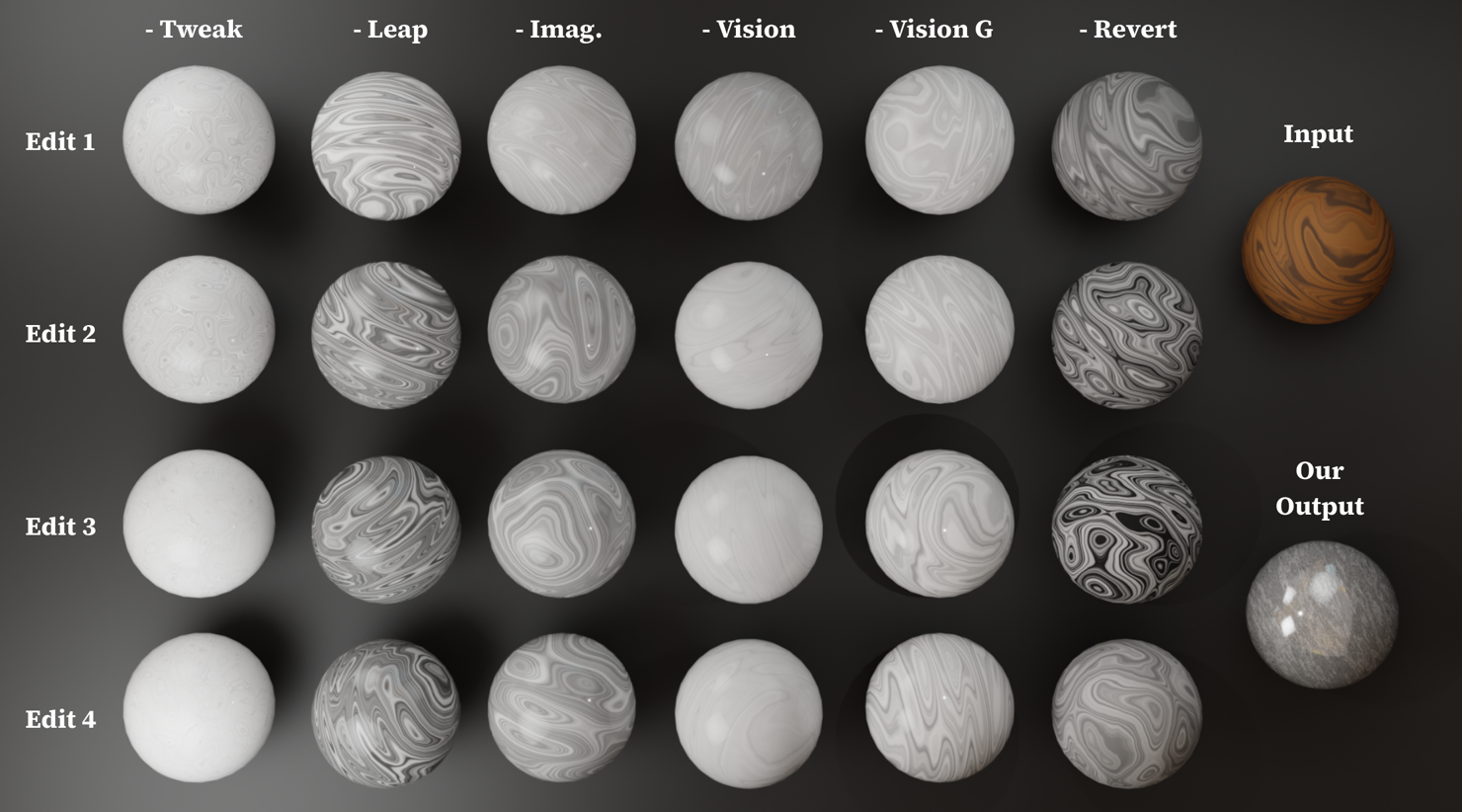}
    
    \caption{\textbf{Qualitative samples of outputs at every editing step of a $4 \times 8$ version of BlenderAlchemy, across different ablations.} For reference, we show the input and the output of our unablated system on the right.}
    \label{fig:ablation_examples}
\end{figure}

So far, we've investigated the quantitative effects of ablating various parts of our system on metrics measuring the alignment of a material with a user's intention. We now discuss the qualitative effects of these ablations on the output on the material editing task, with reference to a sample shown in Figure \ref{fig:ablation_examples}. 

\paragraph{\textbf{Tweak and leap edits.}} The columns ``- Tweak'' and ``- Leap'' correspond to leap-only and tweak-only versions of BlenderAlchemy. When leap edits are disabled (``- leap''), we can see that the edits fail to change the structure of the swirls, but instead produce darker stripes in an attempt to make the output look more marble-like, a change that can be associated with continuous parameters of certain graph nodes. On the other hand, when tweak edits are disabled, BlenderAlchemy produces drastic changes to the swirling patterns of the wood in Edit 1, but leads to a plateauing of progress, as all subsequent edits are drastic enough that reversion reverts back to edit 1, making no progress beyond the pale white material in edit 1.

\paragraph{\textbf{Visual imagination.}} The column ``- Imag.'' corresponds to ablations of visual imagination for the text-based material editing task. Note how though edits are being made in every edit iteration, the verisimilitude of the material plateaus very quickly. Without a visual target to compare against, the edit generator has a difficult time knowing how to adjust the parameters of the shader node graph.

\paragraph{\textbf{Visual perception}} Columns ``-Vision'' and ``- Vision G'' correspond to ablating (1) the vision of the edit generator \textit{and} the visual state evaluator and (2) ablating the vision of the edit generator \textit{only}. In both cases, we see that it's mostly adjusting the color of the input wood material, making it light grey to match the prompt. However, the end result does not look like marbled granite.

\paragraph{\textbf{Edit hypothesis reversion.}} Column ``- Revert'' shows  what happens when edit hypothesis reversion is disabled. As can be seen in Edit 3, the best candidate among the edit hypotheses is chosen to be a material that is \textit{less} similar to granite marble than Edit 1. Edit 4 recuperates a little, but the instability has costed BlenderAlchemy 2 edit cycles, all just to eventually end up with a material that fits the prompt as much as Edit 1. This shows the importance of providing BlenderAlchemy the ability to revert to earlier edit hypotheses.

\begin{figure}[tb]
\centering
\includegraphics[width=\textwidth]{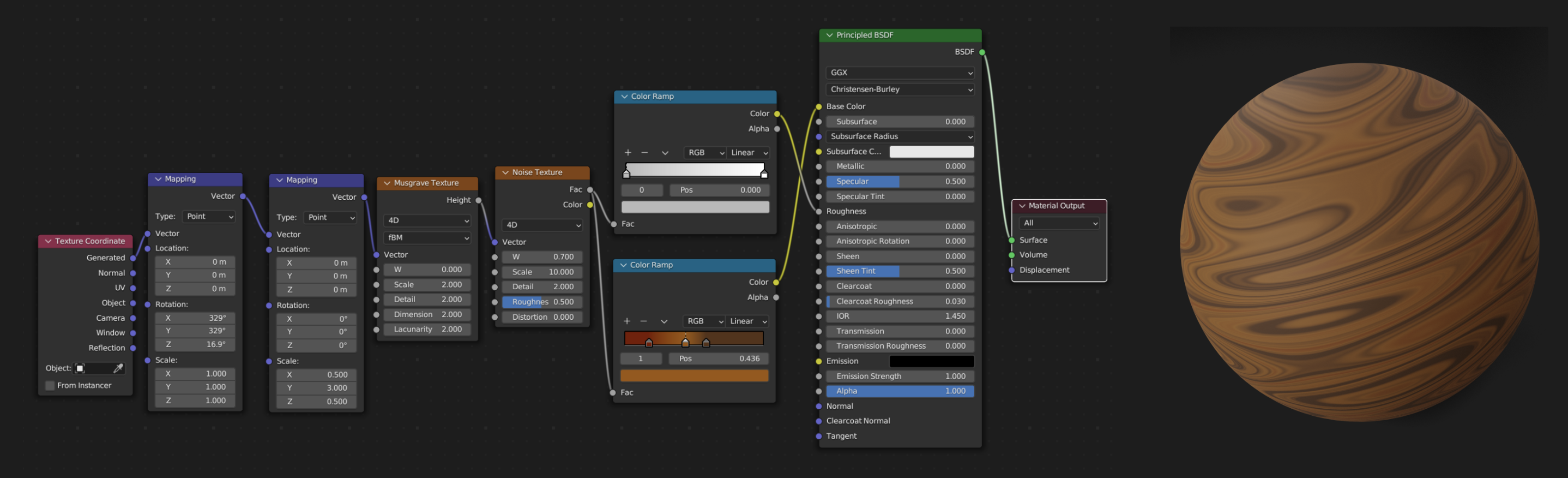}
\caption{\textbf{Blender material graph of the starter wood material.}}
\label{fig:material_graph_wood}
\vspace{-1em}
\end{figure}
\begin{figure}[tb]
\centering
\includegraphics[width=\textwidth]{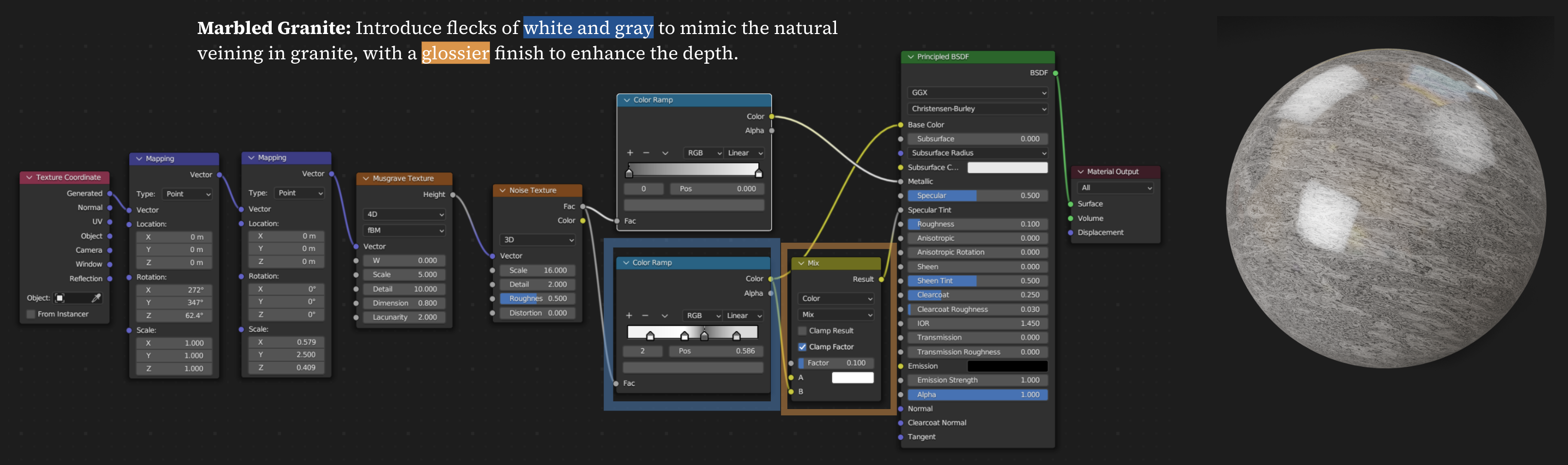}
\caption{\textbf{Blender material graph of the ``marbled 
granite'' material.} Note the correspondence between ``white and gray'' with the colors chosen for the color ramp, and ``glossy finish'' with the input into the specularity port of the principled BSDF node.}
\label{fig:material_graph_marble}
\vspace{-1em}
\end{figure}
\begin{figure}[tb]
\centering
\includegraphics[width=\textwidth]{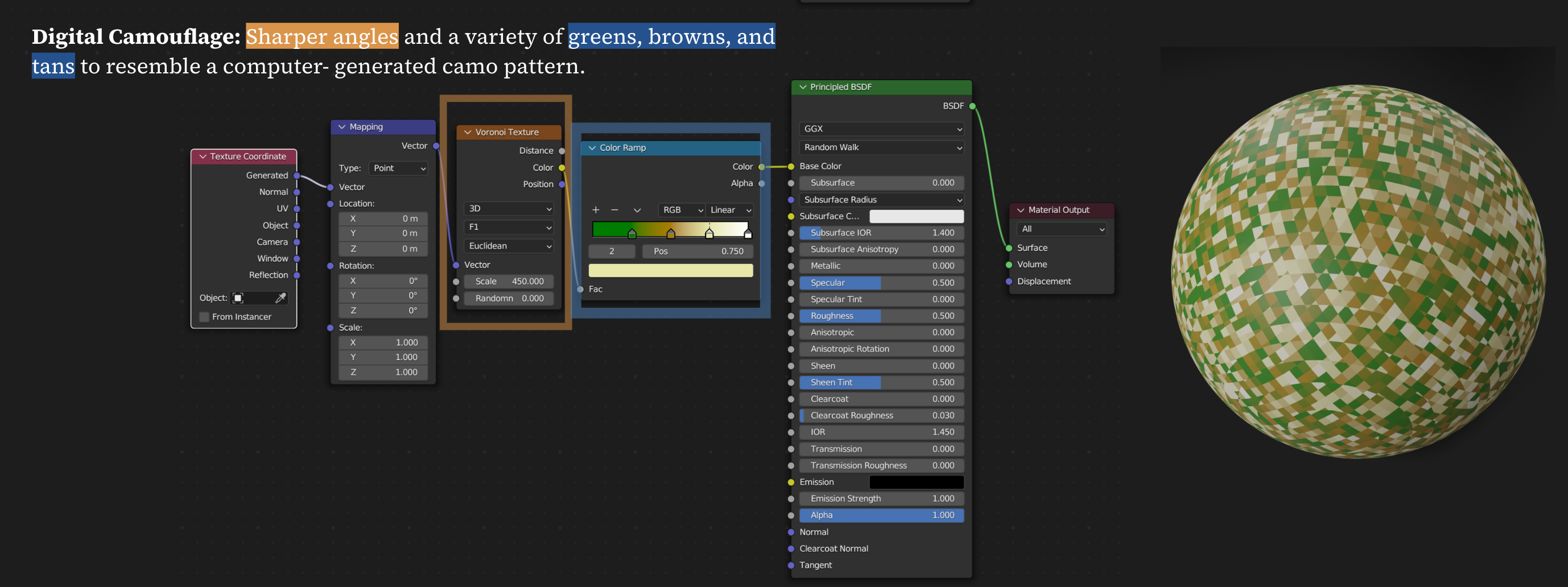}
\caption{\textbf{Blender material graph of the ``digital camo'' material.} To achieve the ``sharp angles'', our system chose to use a Voronoi texture node, and chooses the  right colors in the color ramp to match the ``greens, browns and tans'' mentioned.}
\label{fig:material_graph_camo}
\vspace{-1em}
\end{figure}
\begin{figure}[tb]
\centering
\includegraphics[width=\textwidth]{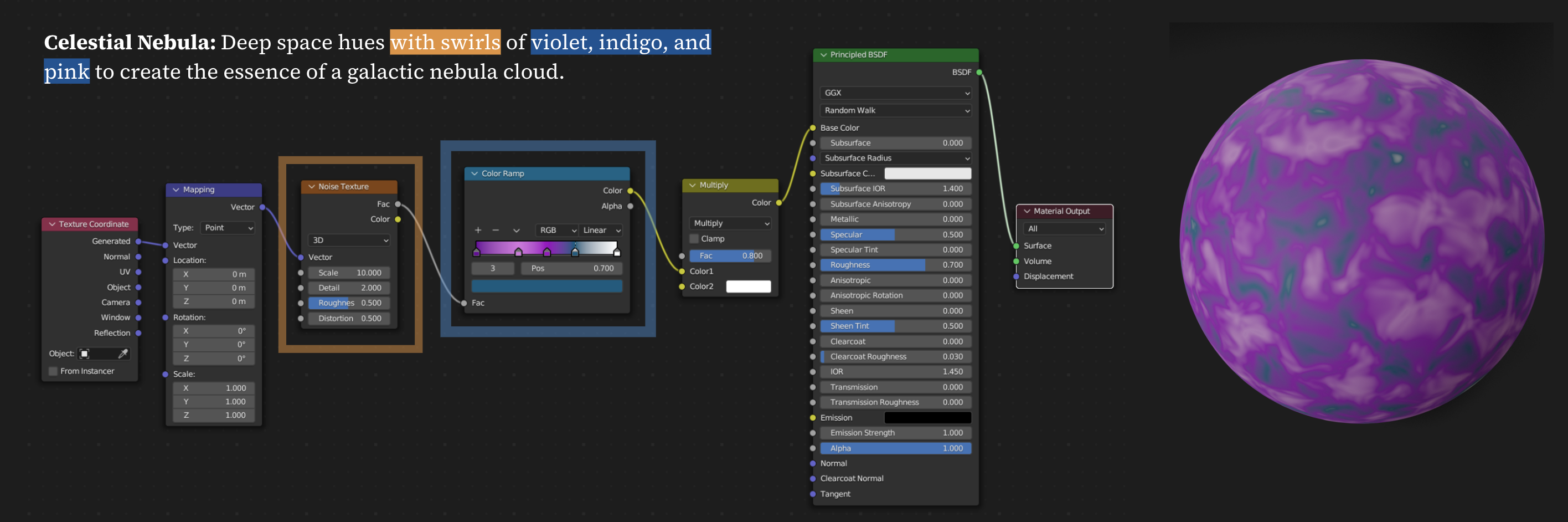}
\caption{\textbf{Blender material graph of the ``celestial nebula'' material.} Note the correspondence between ``swirls'' and the noise texture node, as well as the colors ``violet, indigo and pink'' being reflected in the color ramp.}
\label{fig:material_graph_nebula}
\vspace{-1em}
\end{figure}

\section{Prompting the State Evaluator and the Edit Generator} \label{sec:prompts}

For the material-editing task setting, the prompts used for our edit generator are shown in Figures \ref{fig:prompt_edit_generator_leap} (for leap edits) and \ref{fig:prompt_edit_generator_tweak} (for tweak edits). The prompt used for our visual state evaluator is shown in Figure \ref{fig:prompt_edit_evaluator}. For the ablations of visual perception and visual imagination in our main paper, we adapt each of these prompts. For ablation of visual imagination, we remove the target image from the $G$ prompt and the $V$ prompt, and use the target description instead. An example of this for $V$'s prompt can be seen in Figure \ref{fig:prompt_edit_evaluator_noimag}. For ablation of visual perception of either (or both) $G$ and/or $V$, we modify the prompts for the state evaluator to judge between two candidates based on their code only (Figure \ref{fig:prompt_edit_evaluator_novis}) and do the equivalent with $G$, where we provide only the target description and the source code, and ask it to generate code directly. For the images of materials, we render a $512 \times 512$ image from the same camera in the Blender design space, on one side of a sphere onto which a material is applied. 

The prompts that we use for editing lighting 
configurations are essentially the same as the ones shown for materials, with minor changes to swap mentions of materials  to mentions of lighting setup.

\begin{figure}
    \centering
    \noindent\begin{mdframed}
    \begin{lstlisting}[escapechar=@]
The following Blender code was used to produce a material:
```python 
@\colorbox{light-gray}{[INITIAL MATERIAL CODE]}@
```

The final material is assigned to the object `material_obj`, a sphere, and produces the rendering on the left below:

@\colorbox{light-gray}{[IMAGES OF CURRENT MATERIAL AND TARGET MATERIAL]}@

The desired material is shown in the image on the right. Please describe the difference between the two materials, and edit the code above to reflect this desired change. Pay special attention to the base color of the materials.
MAKE SURE YOUR CODE IS RUNNABLE. MAKE SURE TO ASSIGN THE FINAL MATERIAL TO `material_obj` (through `apply(material_obj)`) AS THE LAST LINE OF YOUR CODE.
DO NOT BE BRIEF IN YOUR CODE. DO NOT ABBREVIATE YOUR CODE WITH "..." -- TYPE OUT EVERYTHING.
    \end{lstlisting}
    \end{mdframed}
    \caption{Prompts used to generate \textbf{leap} edits on an input material.}
    \label{fig:prompt_edit_generator_leap}
\end{figure}
\begin{figure}
    \centering
    \noindent\begin{mdframed}
    \begin{lstlisting}[escapechar=@]
The following Blender code was used to produce a material:
```python 
@\colorbox{light-gray}{[INITIAL MATERIAL CODE]}@
```

The final material is assigned to the object `material_obj`, a sphere, and produces the rendering on the left below:

@\colorbox{light-gray}{[IMAGES OF CURRENT MATERIAL AND TARGET MATERIAL]}@

The desired material is shown in the image on the right. 
Answer the following questions:
1) What is the SINGLE most visually obvious difference between the two materials in the image above?
2) Look at the code. Which fields/variables which are set to numerical values are most likely responsible for the obvious visual difference in your answer to question 1?
3) Copy the code above (COPY ALL OF IT) and replace the assignments of such fields/variables accordingly!
MAKE SURE YOUR CODE IS RUNNABLE. MAKE SURE TO ASSIGN THE FINAL MATERIAL TO `material_obj` (through `apply(material_obj)`) AS THE LAST LINE OF YOUR CODE.
DO NOT BE BRIEF IN YOUR CODE. DO NOT ABBREVIATE YOUR CODE WITH "..." -- TYPE OUT EVERYTHING.
    \end{lstlisting}
    \end{mdframed}
    \caption{Prompts used to generate \textbf{tweak} edits on an input material.}
    \label{fig:prompt_edit_generator_tweak}
\end{figure}
\begin{figure}
    \centering
    \noindent\begin{mdframed}
    \begin{lstlisting}[escapechar=@]
Here is the target material rendering:
@\colorbox{light-gray}{[IMAGE OF TARGET MATERIAL]}@

Below, I show two different materials. Which one is visually more similar to the target material rendering? The one on the left or right?

@\colorbox{light-gray}{[CONCATENATED IMAGES OF 2 CANDIDATES]}@

    \end{lstlisting}
    \end{mdframed}
    \caption{Prompts used to evaluate visual state of the edited material.}
    \label{fig:prompt_edit_evaluator}
\end{figure}

\begin{figure}
    \centering
    \noindent\begin{mdframed}
    \begin{lstlisting}[escapechar=@]
Our desired target material can be described by: @\colorbox{light-gray}{[TARGET DESCRIPTION]}@.
Imagine I'm showing you two Blender python scripts for materials, and they're side by side. Which one has the highest chance of producing the desired target material in Blender? The one on the left or right?
Code on the LEFT:
```python
@\colorbox{light-gray}{[CODE OF CANDIDATE 1]}@
```
Code on the RIGHT:
```python
@\colorbox{light-gray}{[CODE OF CANDIDATE 2]}@
```
Make sure that your final answer indicates which one has the highest chance of producing the desired material -- left or right. Answer by putting left or right in ```'s.

   \end{lstlisting}
    \end{mdframed}
    \caption{Prompts used to evaluate visual state of the edited material, \textbf{when used without vision}.}
    \label{fig:prompt_edit_evaluator_novis}
\end{figure}
\begin{figure}
    \centering
    \noindent\begin{mdframed}
    \begin{lstlisting}[escapechar=@]
Our desired target material can be described by: @\colorbox{light-gray}{[TARGET DESCRIPTION]}@.
Below, I show two different materials. Which one is visually more similar to the desired material described? The one on the left or right?
@\colorbox{light-gray}{[CONCATENATED IMAGES OF 2 CANDIDATES]}@
    \end{lstlisting}
    \end{mdframed}
    \caption{Prompts used to evaluate visual state of the edited material, \textbf{when used without visual imagination for the target material}.}
    \label{fig:prompt_edit_evaluator_noimag}
\end{figure}

\section{Analysis of Program Edits} \label{sec:code_analysis}

What kinds of qualitative changes to programs and material ndoe graphs does BlenderAlchemy affect? Figure \ref{fig:material_graph_wood} shows the wood material shader node graph corresponding to the input material to the text-based material editing task setting. As a few samples, Figures \ref{fig:material_graph_marble}, \ref{fig:material_graph_nebula}, \ref{fig:material_graph_camo} show the material shader node graph of 3 materials edited from the input material. 

Starting with the wood shader graph, one can observe changes of graph node types (e.g. insertion of the Voronoi node in camo material) or connectivity (e.g. changing the ports of the Principled BSDF node that receive inputs from the color ramps in the marbled granite material), continuous values (e.g. color values of the color ramps of the celestial nebula material) or categorical values in nodes (e.g. changing  4D to 3D noise textures in the marbled granite material).

\begin{figure}
    \centering
    \begin{subfigure}[b]{0.6\textwidth}
        \centering
        \includegraphics[width=\textwidth]{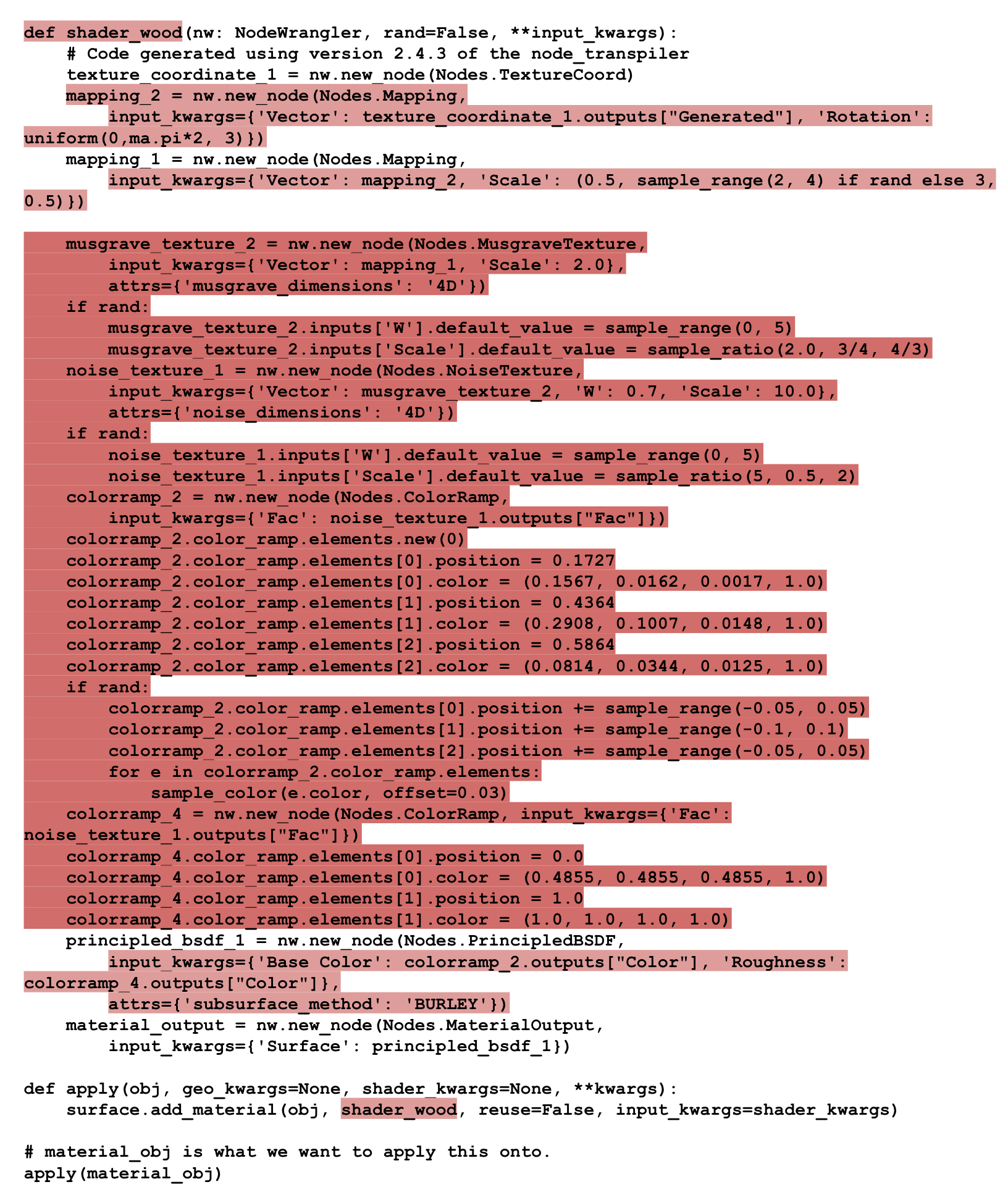}
        \caption{Code for the input synthetic wood material.}
        \label{fig:camo_code_before}
    \end{subfigure}
    \hfill
    \begin{subfigure}[b]{0.6\textwidth}
        \centering
        \includegraphics[width=\textwidth]{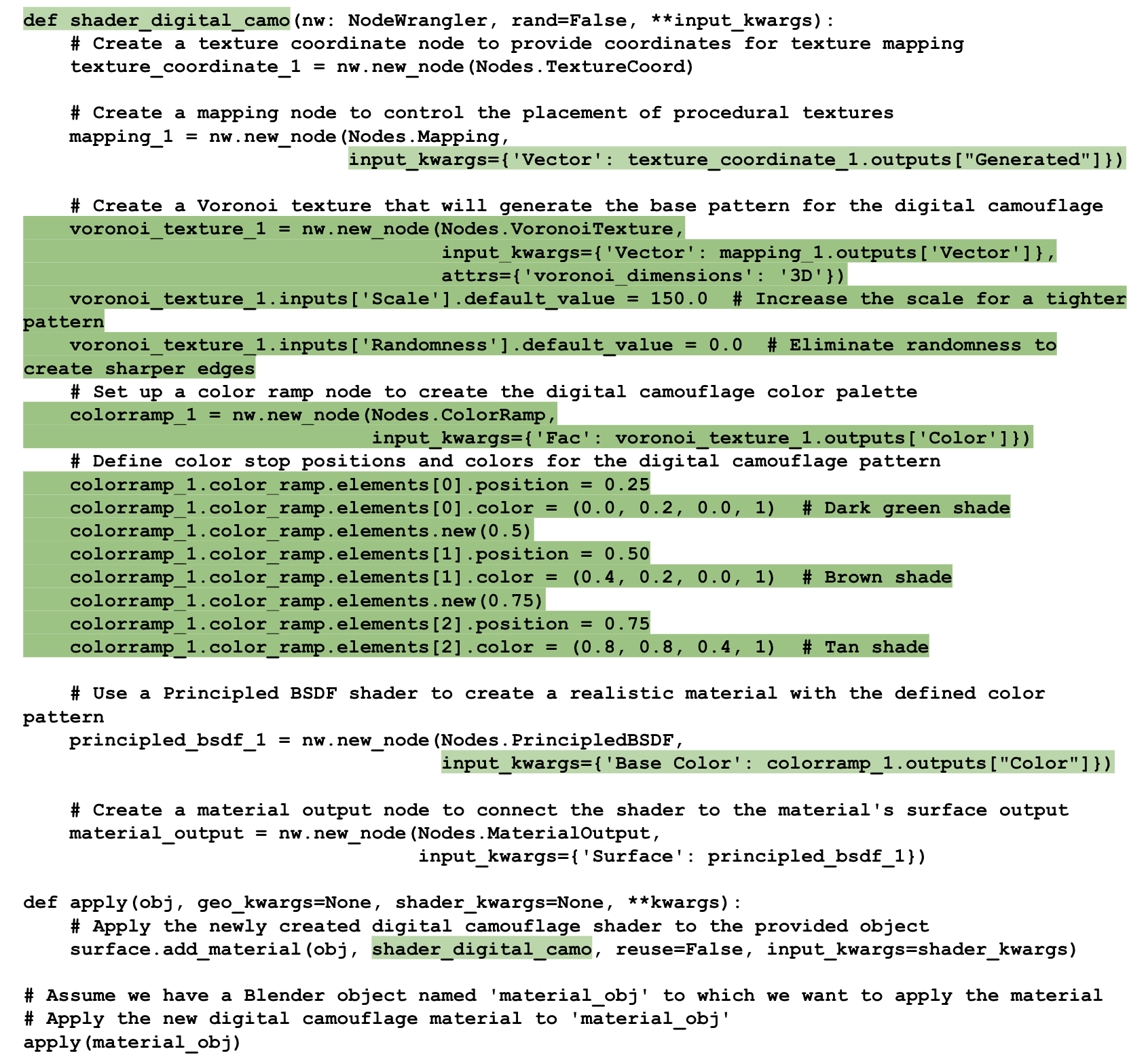}
        \caption{Final output from BlenderAlchemy given the ``digital camo'' prompt.}
        \label{fig:camo_code_after}
    \end{subfigure}
    \caption{The input and output code for the ``digital camouflage'' example. Red parts correspond to parts that are deleted, and green corresponds to parts added by BlenderAlchemy. The darker red/green correspond to the editing of the bulk of the procedural node setup for both materials. Import statements omitted.}
    \label{fig:camo_code_edit}
\end{figure}

We can observe this in the code as well. An example an be seen in the code edits for the digital camouflage material, in Figure \ref{fig:camo_code_edit}. Figure \ref{fig:camo_code_before} shows the input Blender script of the wooden material (whose graph and rendering are shown in Figure \ref{fig:material_graph_wood}). BlenderAlchemy eventually edits the material by outputting the code in \ref{fig:camo_code_after}. We see that not only does the code replace many lines, eliminating numerous nodes in the wooden material, but also instantiates new nodes like the Voronoi Texture node with appropriate parameters to match the input text description (e.g. see the line setting \texttt{voronoi\_texture\_1.inputs[`Randomness']}, with the comment, ``\# Eliminate randomness to create sharper edges''). See top  left of Figure \ref{fig:material_graph_camo} for the original text description.

\begin{figure}[tb]
    \centering
    \begin{subfigure}[b]{\textwidth}
        \centering
        \includegraphics[width=\textwidth]{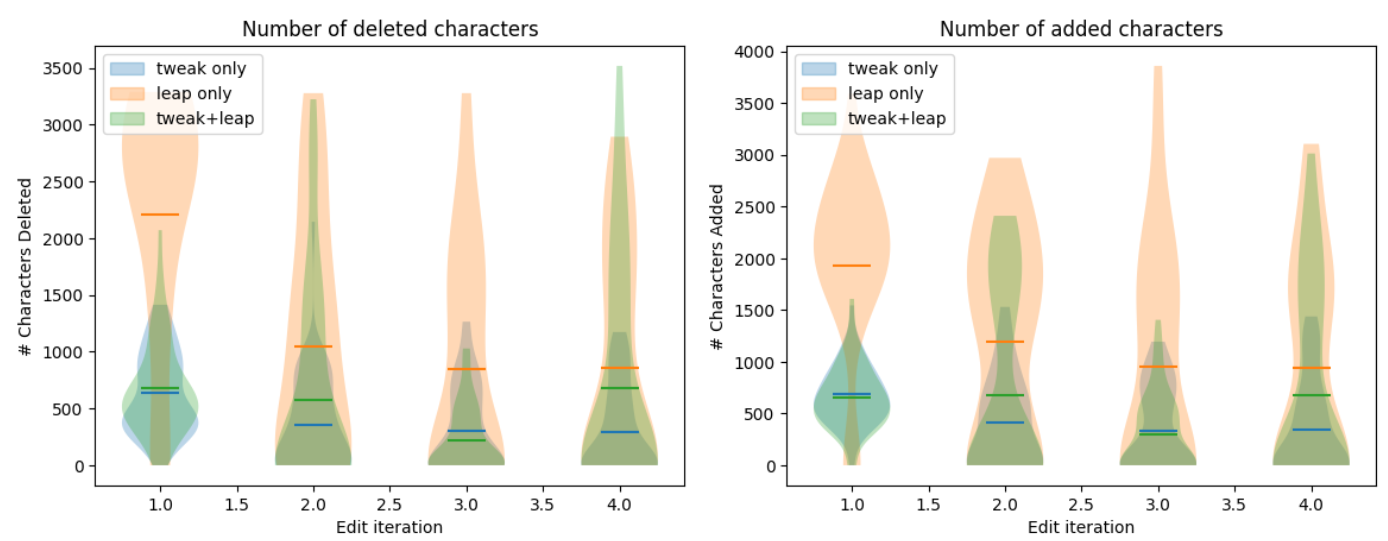}
        \caption{Number of characters added and deleted in editing, across 4 edit iterations.}
        \label{fig:edit_size_numchars}
    \end{subfigure}
    \hfill
    \begin{subfigure}[b]{\textwidth}
        \centering
        \includegraphics[width=\textwidth]{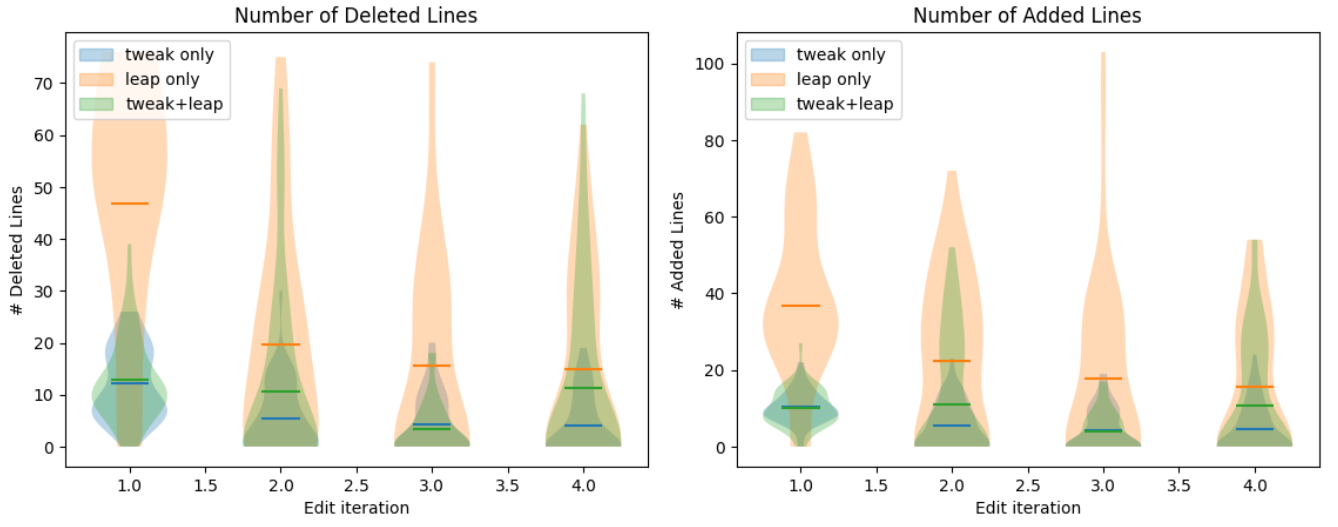}
        \caption{Number of lines added and deleted in editing, across 4 edit iterations.}
        \label{fig:edit_size_numlines}
    \end{subfigure}
    \caption{Analysis of the size of code edits at every step of the editing discovery process of a $4 \times 8$ version of our system. We shown distributions for every edit iteration (1-4) of the number of characters/lines deleted (left) and added (right), for three variants of the system. \textbf{tweak-only} indicates the version where every edit iteration is requested to be tweak edits. \textbf{leap-only} is the equivalent for leap edits. \textbf{tweak+leap} alternates between the two kinds of edits -- edits 1 and 3 are tweak edits, and the 2 and 4 are leap edits. Lines in each distribution indicates the mean.}
    \label{fig:edit_size_analysis}
\end{figure}

Figure \ref{fig:edit_size_analysis} shows the size of the code edits at each iteration of a $4 \times 8$ version of BlenderAlchemy, measured in terms of \textit{the total number of characters added/deleted} (Figure \ref{fig:edit_size_numchars}), and \textit{the number of lines added/deleted} (\ref{fig:edit_size_numlines}). The size of the edit of the best of iteration $i$ is measured with respect to the best of iteration $i-1$, starting at the input script of the wooden material.

We see that even though the restrictions of edits to tweak and leap edits are not strictly enforced in our procedure (and only softly done through VLM prompting), the distribution of edit size measured in both ways suggest that (1) the size of leap edits are substantially larger than those of tweak edits, and  that (2) our method of oscillating between tweak and leap edits (tweaking for iterations 1 and 3, leaping for iterations 2 and 4) allows the distribution of edit size to spend every other iteration looking more similar to either tweak or leap edits. Interestingly, consistently across all the graphs shown, even when our method is producing leap edits, the average size is still lower than those of leap-only edits. We suspect that this is because tweaking in iterations (3 and 4) gets the material closer to the desired outcome, and the need for radical changes is lowered in the 2nd and 4th iterations, compared to the potentially destabilizing effects of leap-only edits.

\section{BlenderAlchemy Materials In Scenes} \label{sec:applied_materials}

\begin{figure}[tb]
    \centering
    \begin{subfigure}[b]{\textwidth}
        \centering
        \includegraphics[width=0.7\textwidth]{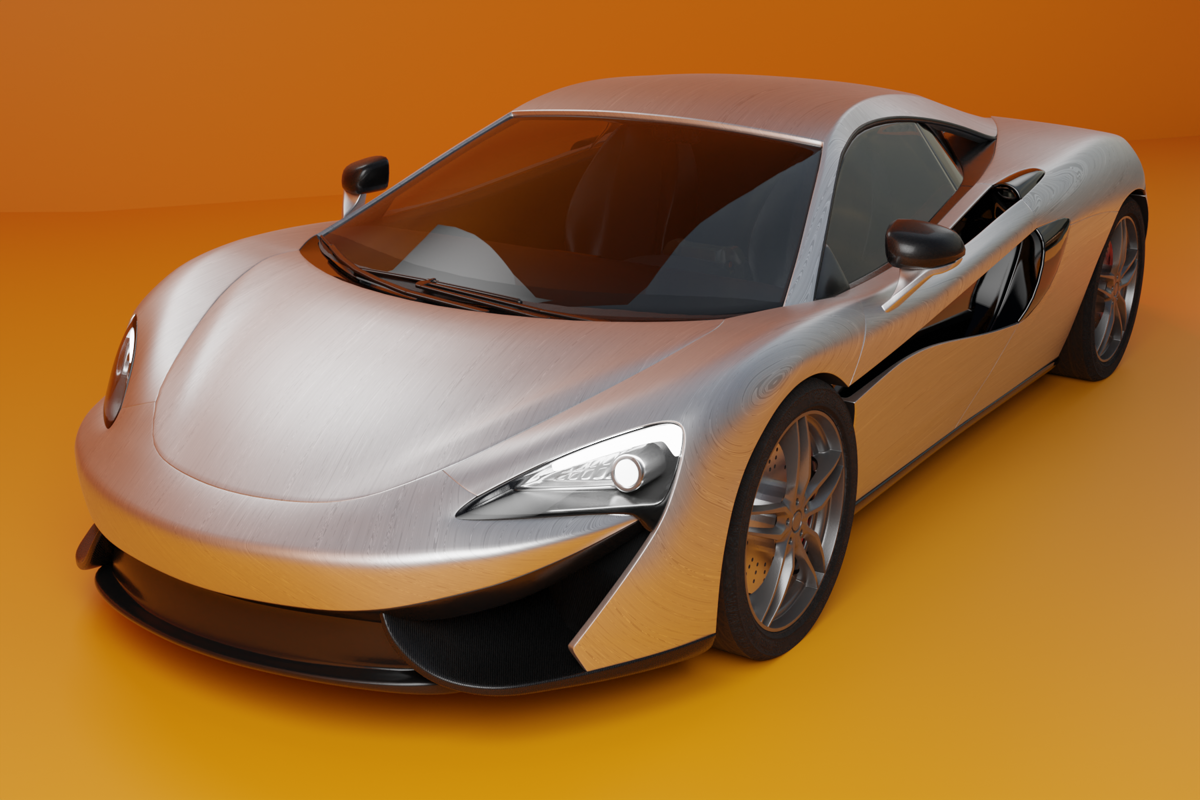}
        \caption{Front view}
    \end{subfigure}
    \hfill
    \begin{subfigure}[b]{\textwidth}
        \centering
        \includegraphics[width=0.7\textwidth]{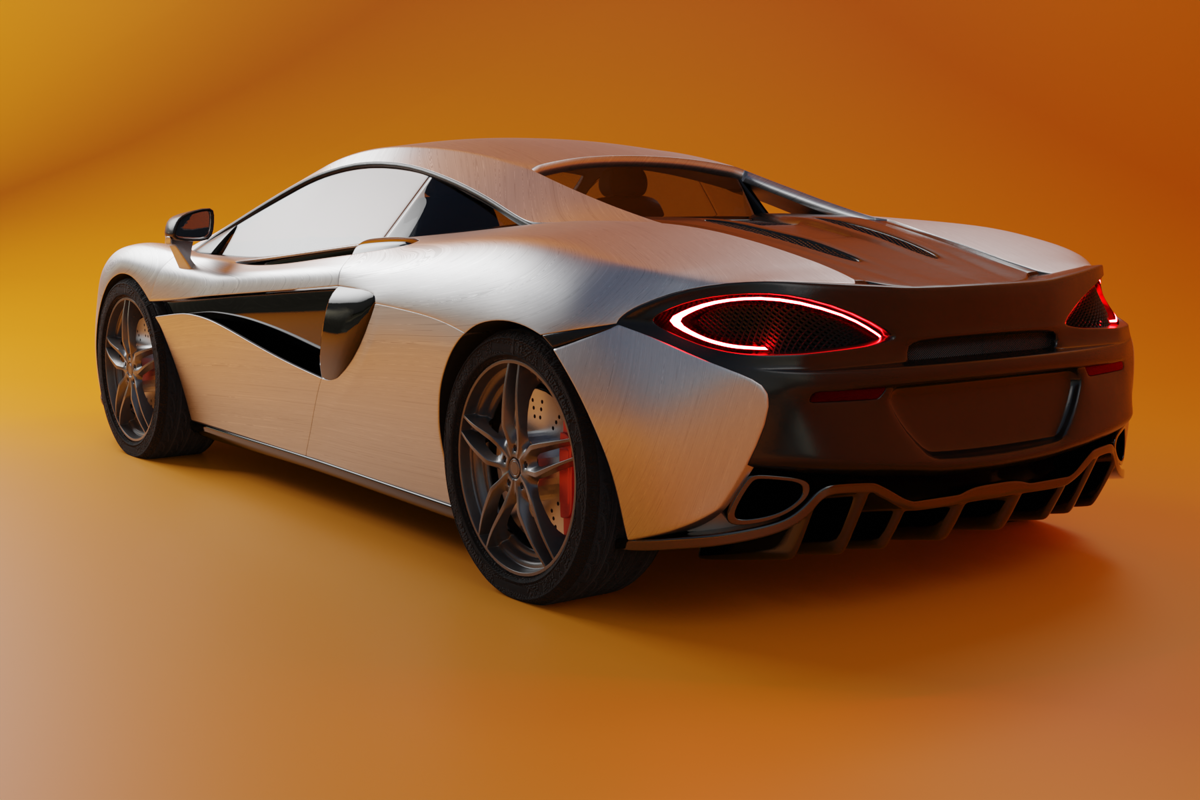}
        \caption{Back view}
    \end{subfigure}
    \hfill
    \begin{subfigure}[b]{0.7\textwidth}
        \centering
        \includegraphics[width=\textwidth]{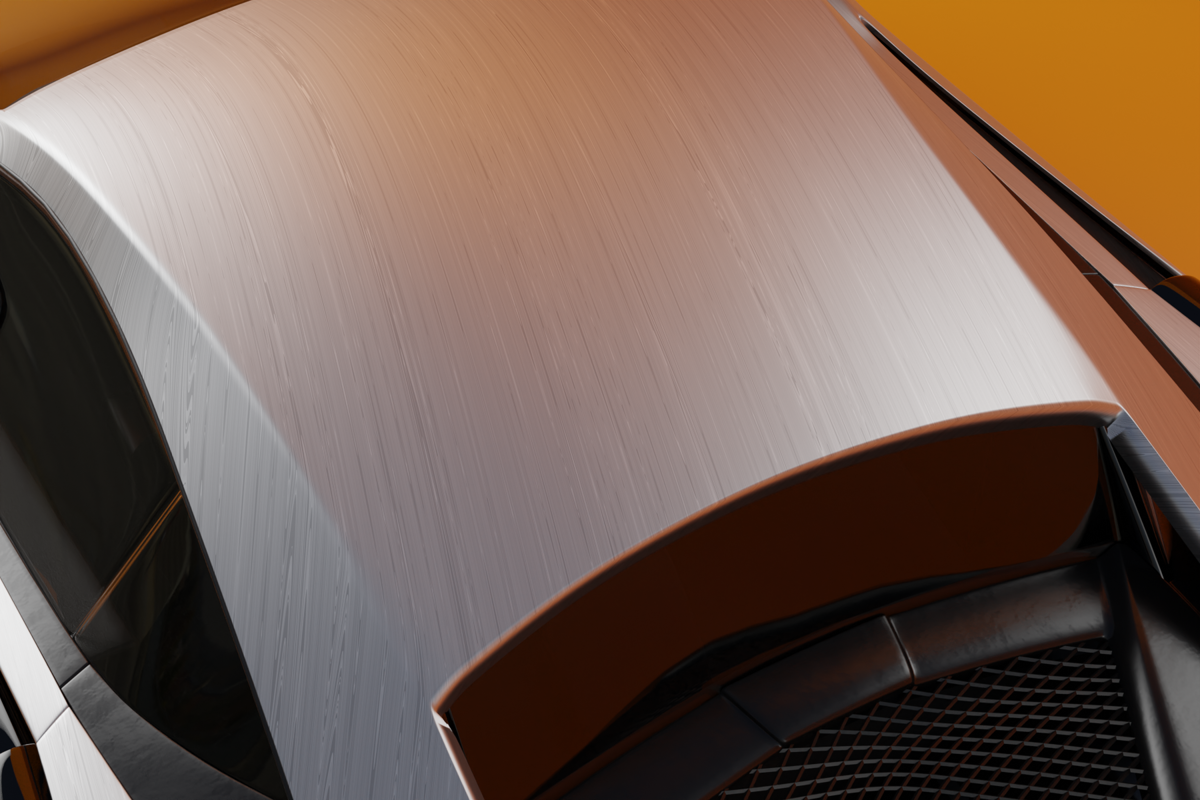}
        \caption{Close up view of brushed aluminum material.}
    \end{subfigure}
    \caption{Brushed aluminum material synthesized by BlenderAlchemy, applied onto the body of a car.}
    \label{fig:fun_car_brushed_aluminum}
\end{figure}

\begin{figure}[tb]
    \centering
    \begin{subfigure}[b]{\textwidth}
        \centering
        \includegraphics[width=0.7\textwidth]{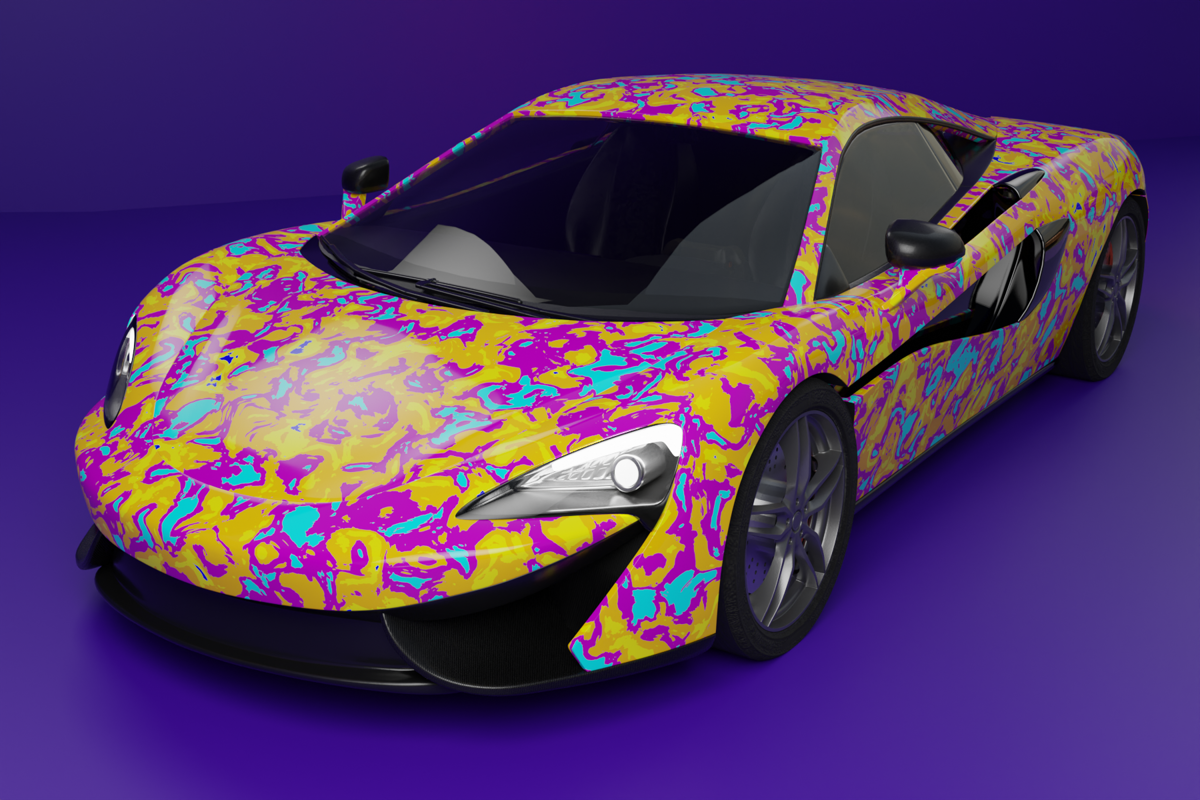}
        \caption{Front view}
    \end{subfigure}
    \hfill
    \begin{subfigure}[b]{\textwidth}
        \centering
        \includegraphics[width=0.7\textwidth]{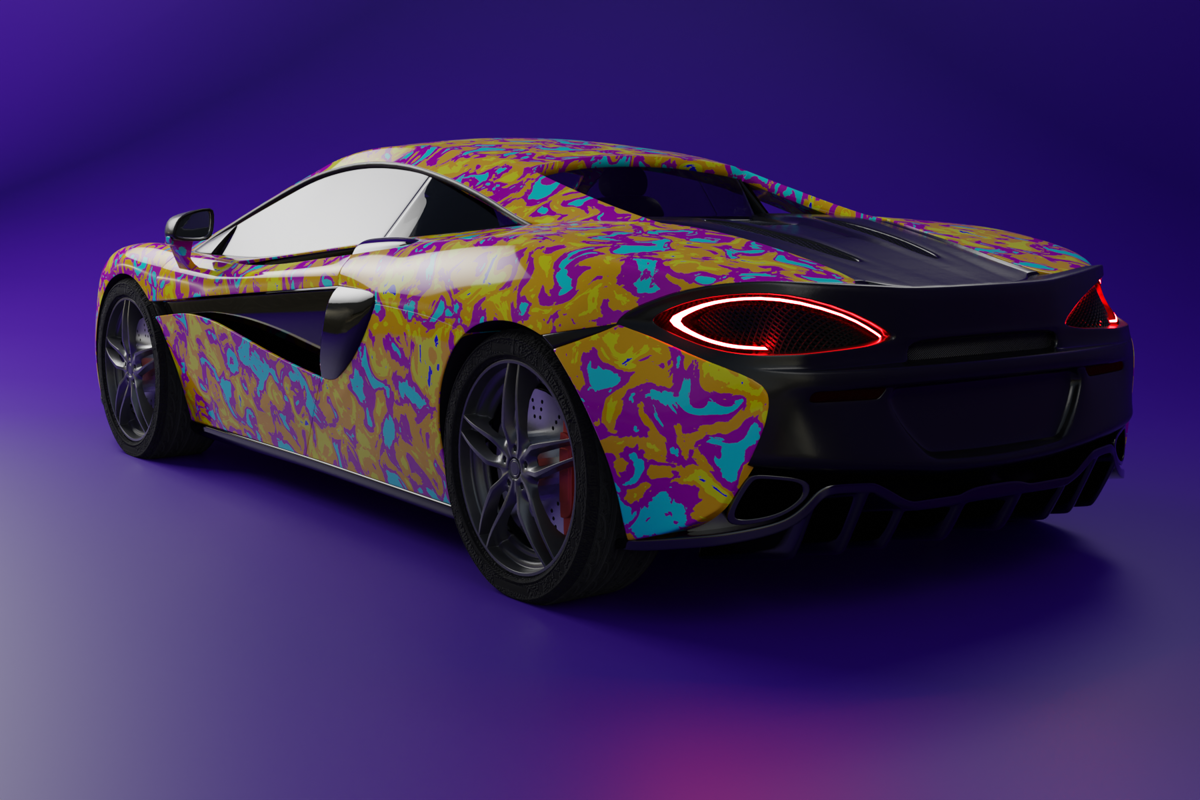}
        \caption{Back view}
    \end{subfigure}
    \hfill
    \begin{subfigure}[b]{0.7\textwidth}
        \centering
        \includegraphics[width=\textwidth]{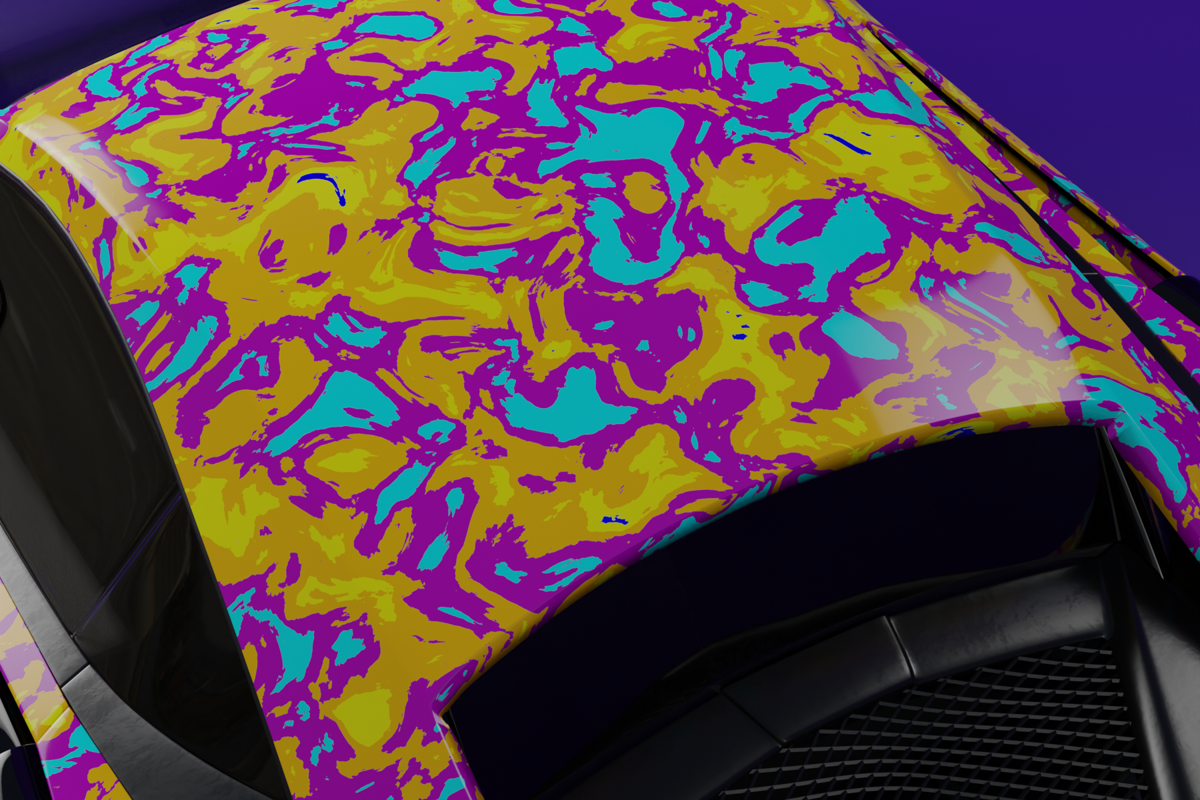}
        \caption{Close up view of the ``paint splash'' material.}
    \end{subfigure}
    \caption{``Paint splash'' material synthesized by BlenderAlchemy, applied onto the body of a car.}
    \label{fig:fun_car_paint}
\end{figure}

\begin{figure}[tb]
    \centering
    \begin{subfigure}[b]{\textwidth}
        \centering
        \includegraphics[width=0.7\textwidth]{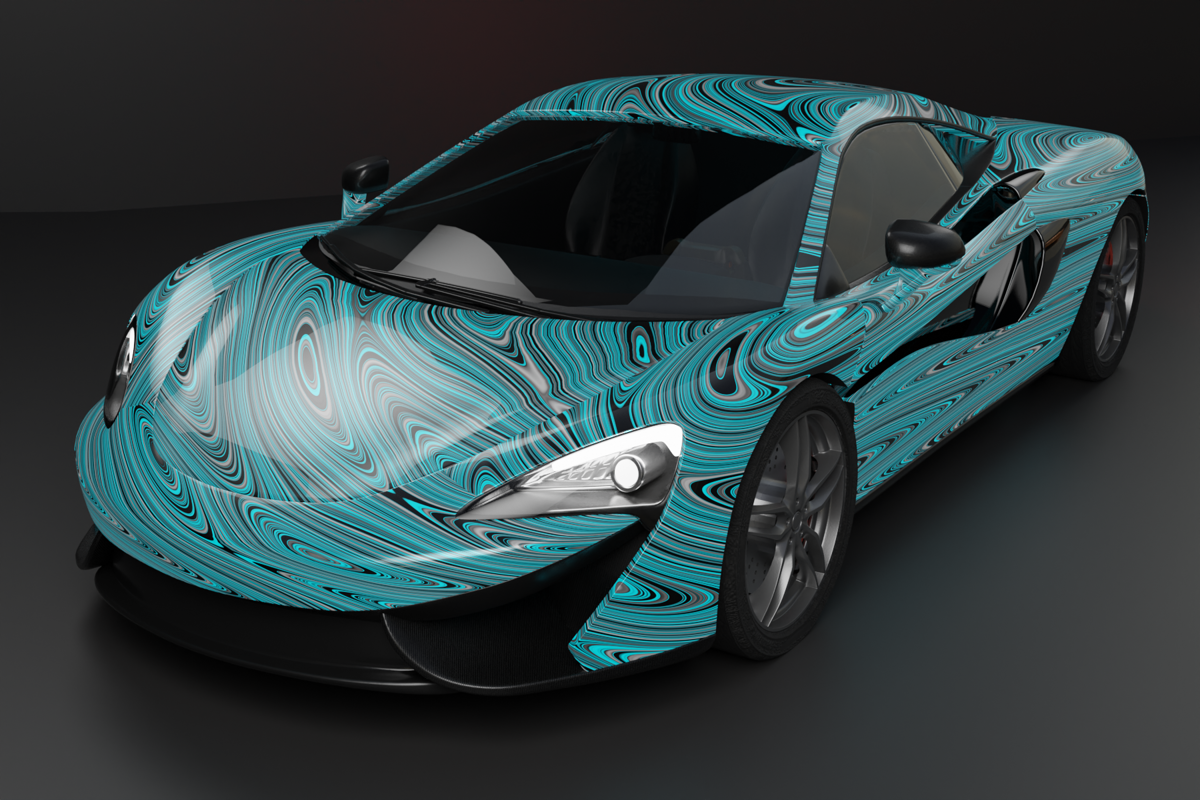}
        \caption{Front view}
    \end{subfigure}
    \hfill
    \begin{subfigure}[b]{\textwidth}
        \centering
        \includegraphics[width=0.7\textwidth]{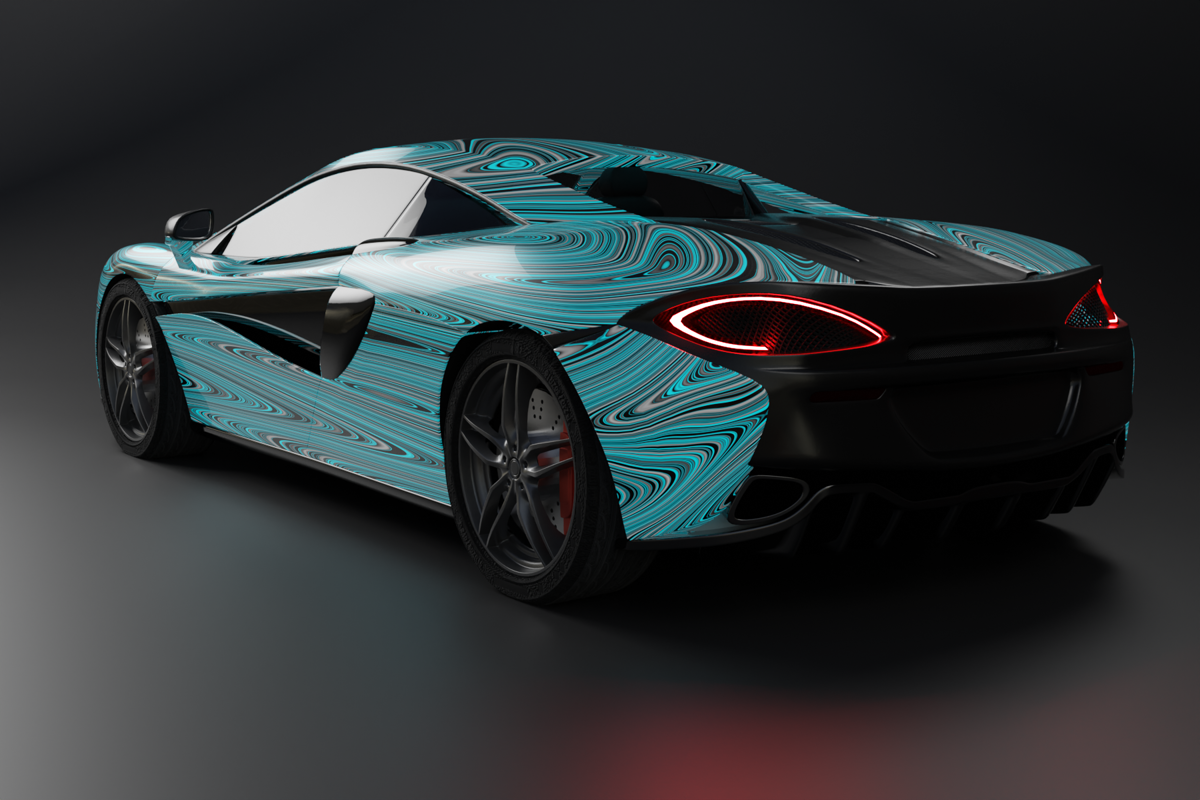}
        \caption{Back view}
    \end{subfigure}
    \hfill
    \begin{subfigure}[b]{0.7\textwidth}
        \centering
        \includegraphics[width=\textwidth]{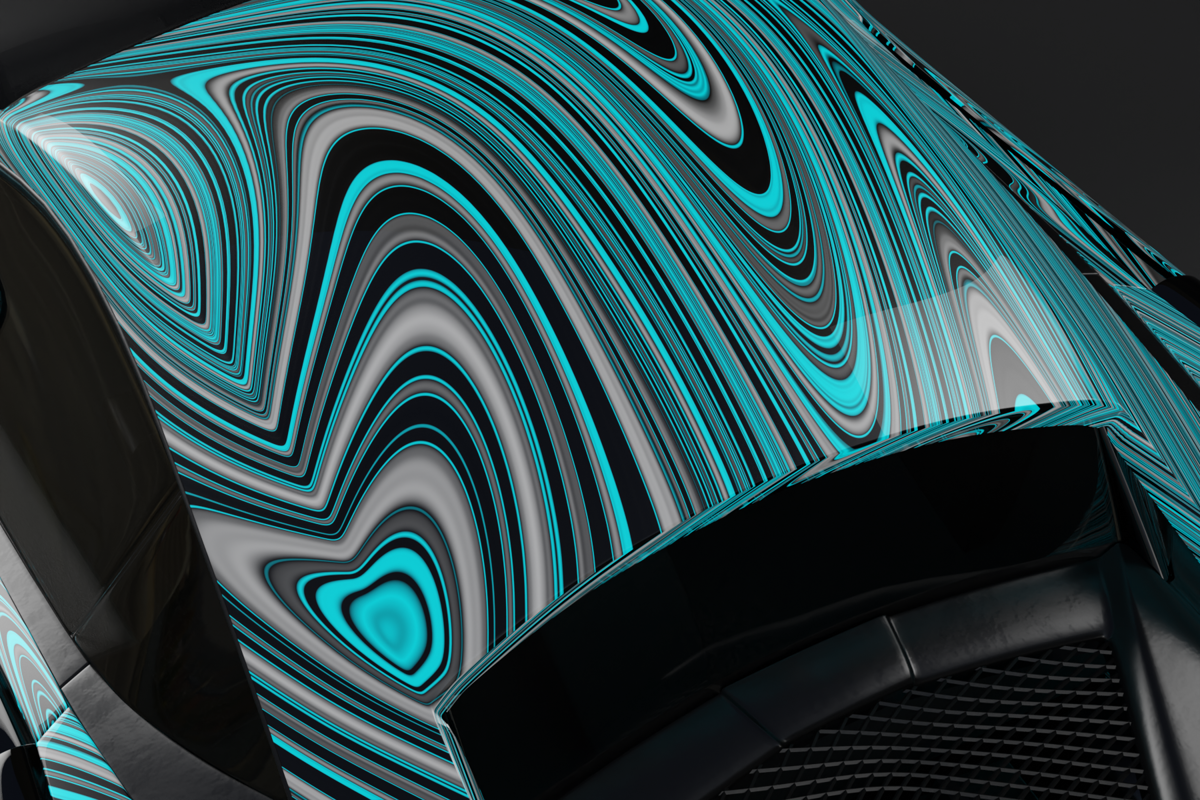}
        \caption{Close up view of the ``Tron Legacy'' material.}
    \end{subfigure}
    \caption{"Tron Legacy" material synthesized by BlenderAlchemy, applied onto the body of a car.}
    \label{fig:fun_car_tron}
\end{figure}

\begin{figure}[tb]
    \centering
    \begin{subfigure}[b]{0.4\textwidth}
        \centering
        \includegraphics[width=\textwidth]{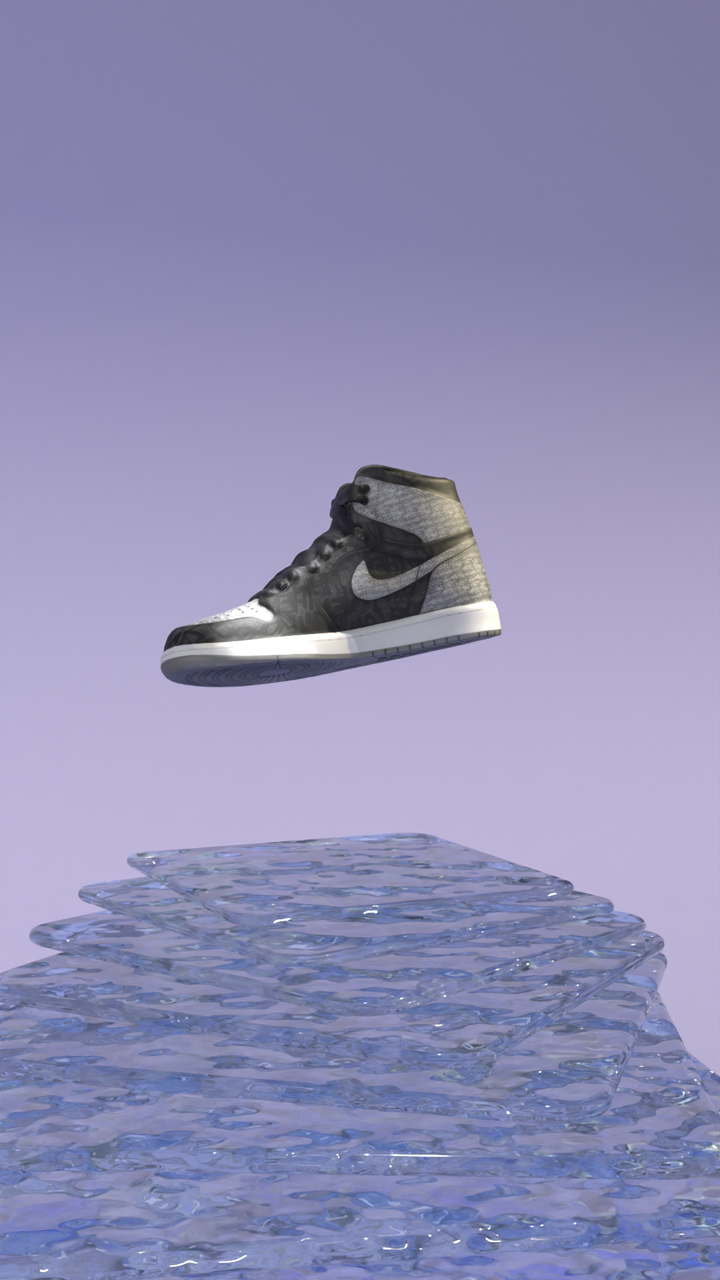}
        \caption{View of whole scene}
    \end{subfigure}
    \begin{subfigure}[b]{0.4\textwidth}
        \centering
        \includegraphics[width=\textwidth]{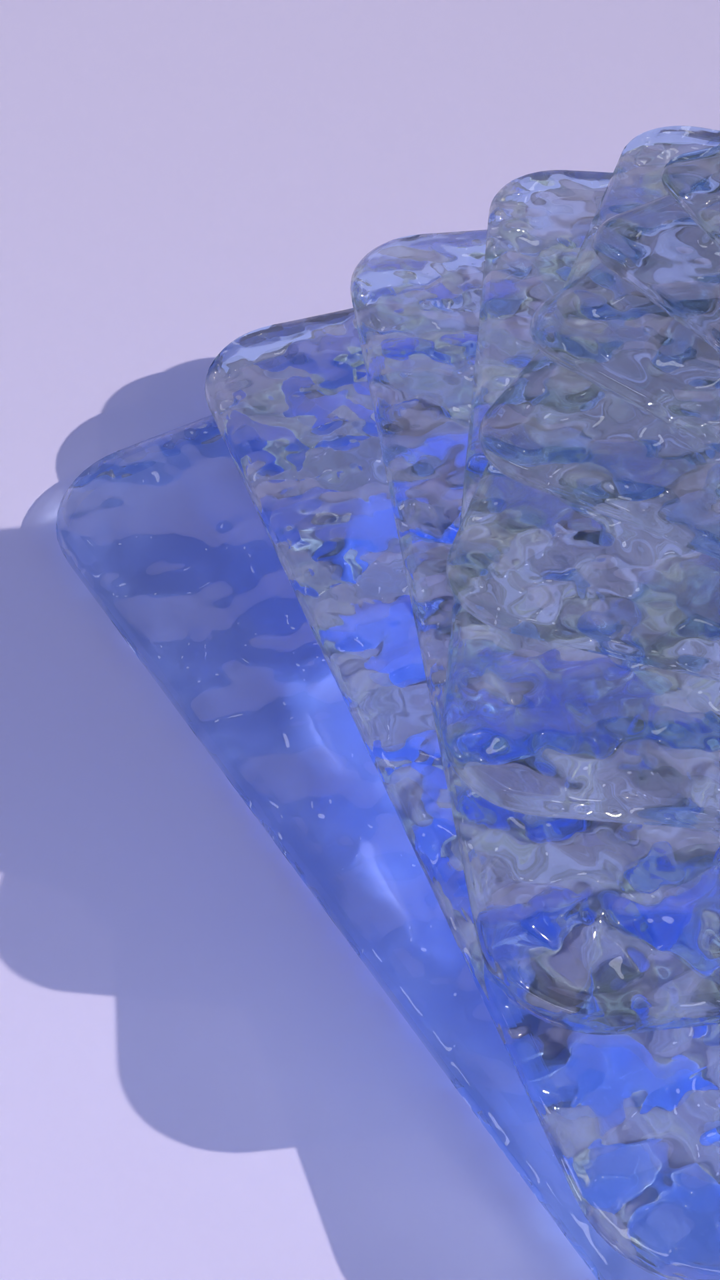}
        \caption{Closeup view of the ice material.}
    \end{subfigure}
    \hfill
    \caption{``Ice'' material synthesized by BlenderAlchemy, applied onto the slabs supporting the shoe.}
    \label{fig:fun_shoe_ice}
\end{figure}

\begin{figure}[tb]
    \centering
    \begin{subfigure}[b]{0.4\textwidth}
        \centering
        \includegraphics[width=\textwidth]{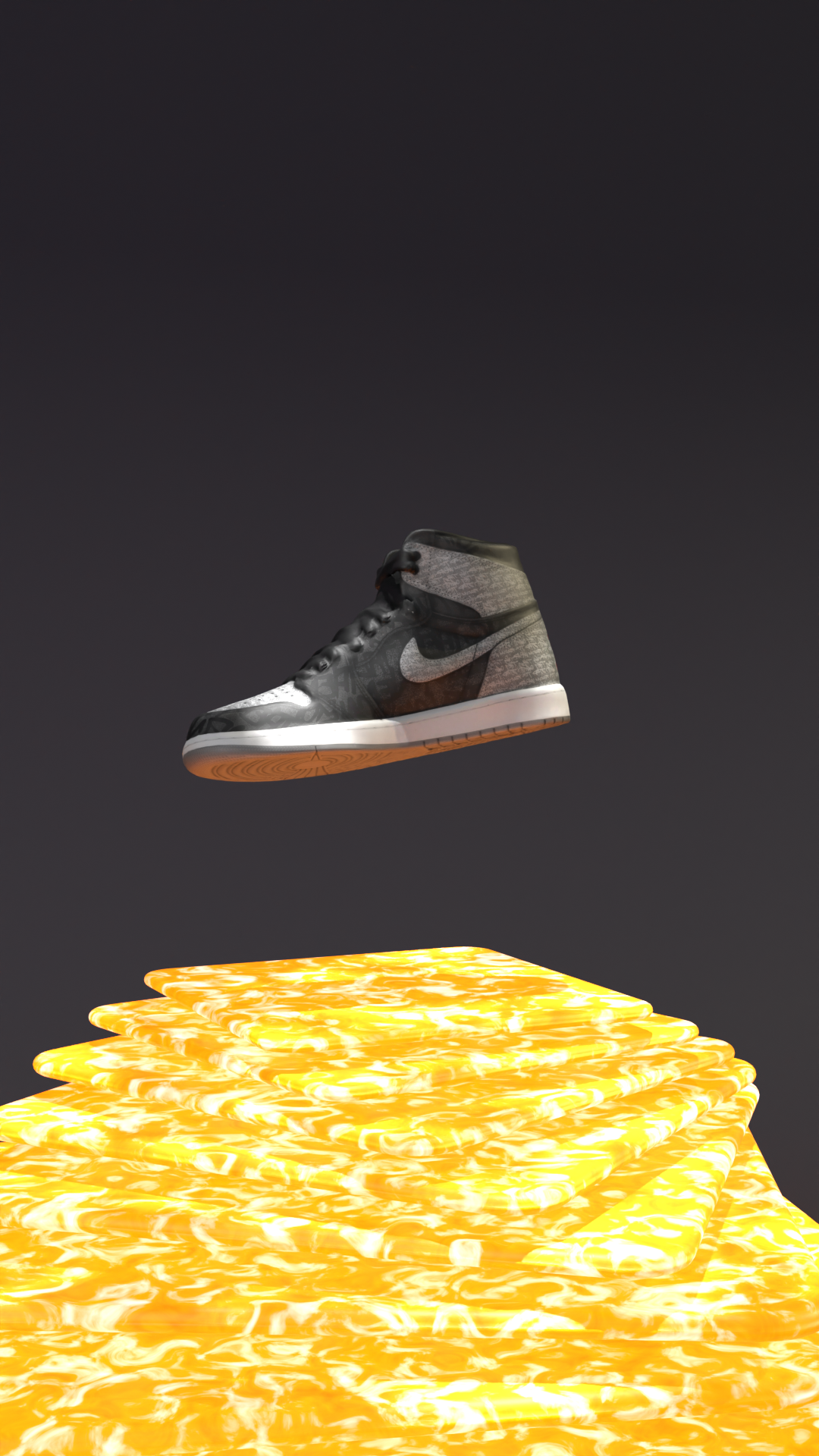}
        \caption{View of whole scene}
    \end{subfigure}
    \begin{subfigure}[b]{0.4\textwidth}
        \centering
        \includegraphics[width=\textwidth]{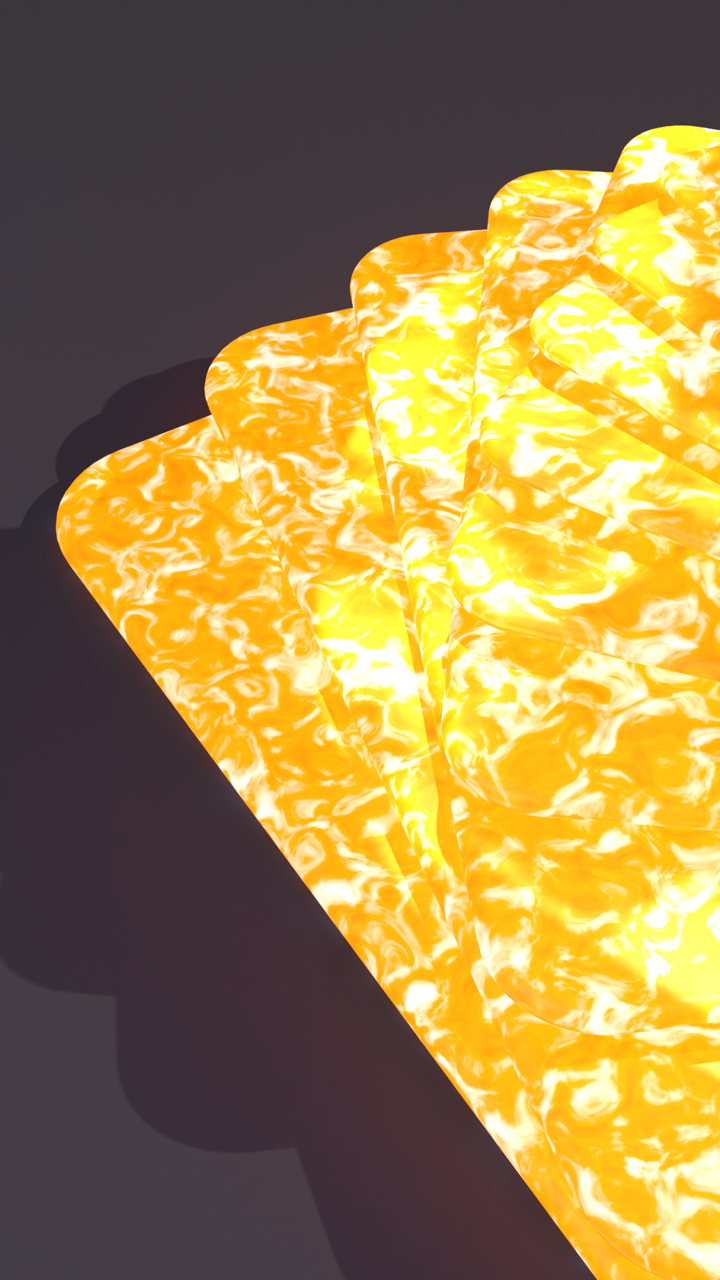}
        \caption{The ``surface of the sun'' material.}
    \end{subfigure}
    \hfill
    \caption{``Surface of the sun'' material synthesized by BlenderAlchemy, applied onto the slabs supporting the shoe.}
    \label{fig:fun_shoe_sun}
\end{figure}

\begin{figure}[tb]
    \centering
    \begin{subfigure}[b]{\textwidth}
        \centering
        \includegraphics[width=\textwidth]{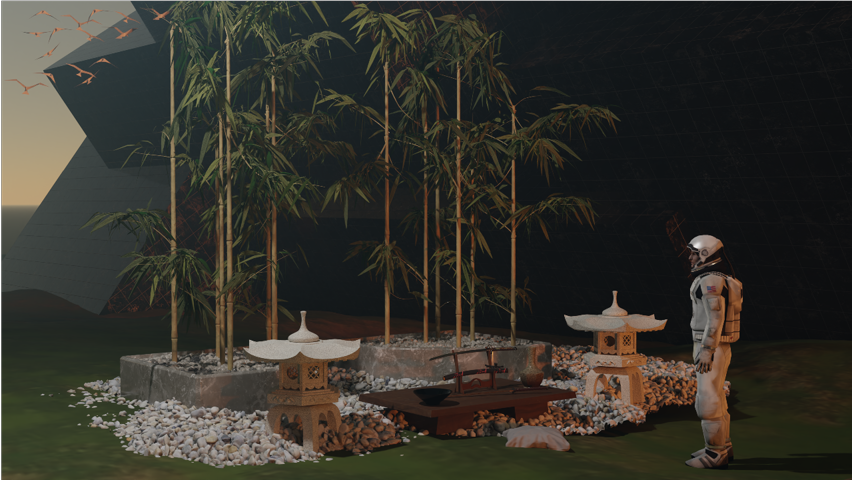}
        \caption{View of whole scene}
    \end{subfigure}
    \begin{subfigure}[b]{\textwidth}
        \centering
        \includegraphics[width=\textwidth]{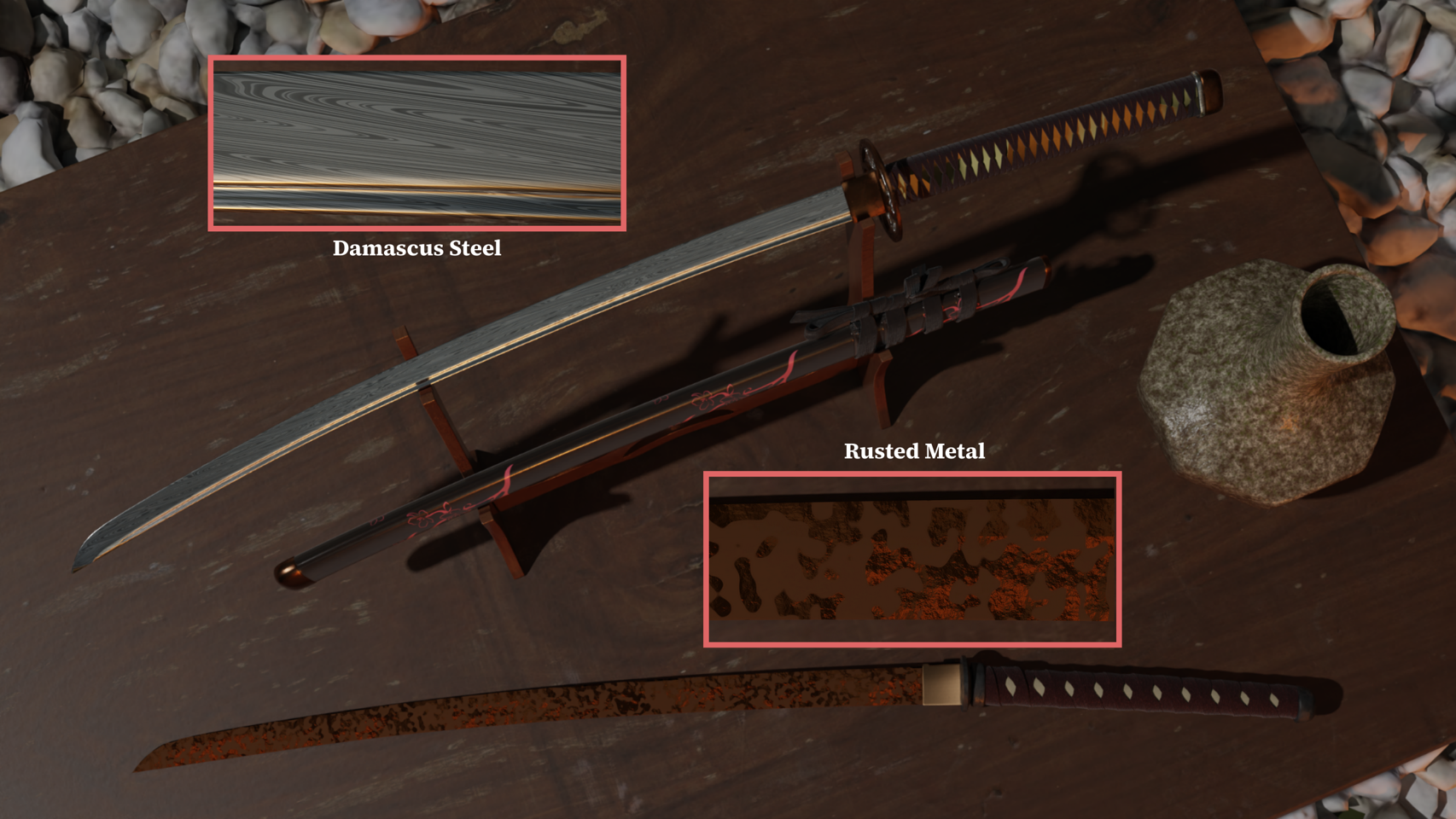}
        \caption{Close up of materials}
    \end{subfigure}
    \hfill
    \caption{Damascus steel and rusted metal synthesized by BlenderAlchemy, applied onto the blades of two katanas in the scene.}
    \label{fig:fun_katana}
\end{figure}

In this section we show the results of applying the material outputs of BlenderAlchemy onto meshes that we download from the internet. Skip to Figures \ref{fig:fun_car_brushed_aluminum},\ref{fig:fun_car_paint},\ref{fig:fun_car_tron},\ref{fig:fun_shoe_ice},\ref{fig:fun_shoe_sun}, \ref{fig:fun_katana} to see the final visualizations. We start with a concise language description, like ``demascus steel'', and then expand the description into a more detailed one by prompting GPT4, using the prompt in Figure \ref{fig:semantic_upsample}, following the semantic-upsampling idea in \cite{huang2023aladdin}.

\begin{figure}
    \centering
    \noindent\begin{mdframed}
    \begin{lstlisting}[escapechar=@]
Come up with an in-detail caption of a material, describing the details of its appearance, including colors, textures, surface characteristics. For instance, for "Marbled Granite", output something like "Marbled Granite: Introduce flecks of white and gray to mimic the natural veining in granite, with a glossier finish to enhance the depth." Now do this for @\colorbox{light-gray}{[INPUT DESCRIPTION]}@. Write no more than 1 sentence.
    \end{lstlisting}
    \end{mdframed}
    \caption{Prompt used to derive more detailed description of the appearance of a material, given an abstract short description.}
    \label{fig:semantic_upsample}
\end{figure}

As the outcome of this, here's the list of (expanded) language descriptions that we use to synthesize the materials for this section:
\begin{enumerate}
    \item Damascus Steel: Swirls of contrasting steely grays and blacks, interwoven to create a mesmerizing, wavelike pattern on the metal's surface, exhibiting a unique blend of toughness and flexibility with a semi-matte finish.
    \item Brushed Aluminum: Present a sleek, matte finish with fine, unidirectional satin lines, exuding an industrial elegance in shades of silver that softly diffuse light.
    \item Surface of the Sun: Envision a vibrant palette of fiery oranges, deep reds, and brilliant yellows, swirling and blending in a tumultuous dance, with occasional brilliant white flares erupting across a textured, almost liquid-like surface that seems to pulse with light and heat.
    \item Ice Slabs: Crystal-clear with subtle blue undertones, showcasing intricate patterns of frozen bubbles and fractures that glisten as they catch the light, embodying the serene, raw beauty of nature's artistry.
    \item  Tron Legacy Material: A sleek, electric blue and black surface with a high gloss finish, featuring circuit-like patterns that glow vibrantly against the dark backdrop, evoking the futuristic aesthetic of the Tron digital world.
    \item Paint Splash Material: A vibrant array of multicolored droplets scattered randomly across a stark, matte surface, creating a playful yet chaotic texture that evokes a sense of spontaneity and artistic expression.
    \item Rusted Metal: A textured blend of deep oranges and browns, with irregular patches and streaks that convey the material's weathered and corroded surface, giving it a rough, tactile feel.
\end{enumerate}

Since BlenderAlchemy requires a starter material script as input, we choose this for each of the above among (1) available procedural materials in Infinigen~\cite{raistrick2023infinite} and (2) materials BlenderAlchemy has synthesized thus far. For the starter materials for each, the following were decided manually:
\begin{enumerate}
    \item  Damascus Steel: We start from the \textbf{metallic  swirl} material (previously synthesized by BlenderAlchemy from an input wood material -- see main paper).
    \item  Brushed Aluminum: We start from the \textbf{metallic swirl} material  (previously synthesized by BlenderAlchemy from an input wood material -- see main paper).
    \item  Surface of the sun: we start from the \textbf{acid trip} material (previously synthesized by BlenderAlchemy from an input wood material -- see main paper).
    \item  Ice Slabs: we start from the \textbf{chunky rock} material from the Infinigen~\cite{raistrick2023infinite} procedural material library. 
    \item Tron Legacy Material: we start from the \textbf{acid trip} material (previously synthesized by BlenderAlchemy from an input wood material -- see main paper).
    \item Paint Splash Material: we start from the \textbf{celestial nebula} material (previously synthesized by BlenderAlchemy from an input wood material -- see main paper).
    \item Rusted Metal : We start from the \textbf{stone} material from the Infinigen~\cite{raistrick2023infinite} procedural material library.
\end{enumerate}

We use a 4x8 version of Blender Alchemy with visual imagination enabled. Look at Figures \ref{fig:fun_car_brushed_aluminum}, \ref{fig:fun_car_paint}, \ref{fig:fun_car_tron} for the application of the brushed aluminum, paint splash and Tron Legacy materials applied to a McLaren. Figures \ref{fig:fun_shoe_ice} and \ref{fig:fun_shoe_sun} show the application of the ice slabs and surface-of-the-sun materials onto a product visualization of a pair of Nikes. Finally, Figure \ref{fig:fun_katana} shows the application of Damascus steel and rusted metal onto two different katanas within an intriguing scene. All scenes are created using assets found on BlenderKit~\cite{blenderkit}.


\section{Societal Impact}\label{sec:impact}

\paragraph{\textbf{Impact on the creative community.}} Blender is the most popular free and open sourced 3D creation suite for both professional and hobbyist 3D designers, and has developed a huge community of artists (\textbf{r/blender} has, as of March 2024, close  to 1.2 million members \cite{blenderreddit}) and developers (the Blender github has 4.1K followers, and 134K commits to their official Blender repo \cite{blendergithub}). Community efforts like Blender open movies ( \url{https://studio.blender.org/films/}) not only demonstrate the versatility and expressive power of the tool itself, but also the collective drive of human imagination. Lowering the barrier to entry for the regular user to participate in such community art projects is the main motivation of BlenderAlchemy. BlenderAlchemy \textit{does not} (and \textit{does not intend to}) replace the intention of the human creator, and is the reason why we've made the starting point of BlenderAlchemy \textit{an existing project file}, and \textit{a specification of intent}. This is in contrast to all the other works like \cite{de2023llmr,blendergpt,sun20233dgpt,yamada2024l3go}, which attempt to generate everything from scratch. The goal of this project is not about the creator \textit{doing} less as much as it is about enabling the creator \textit{to create} more.

\paragraph{\textbf{Systematic Biases.}} LLMs and VLMs have inherent biases that are inherited by BlenderAlchemy. For now, safeguarding against such biases from making it into the final 3D creation requires the careful eye of the human user in the process -- the fact that edits are made directly in Blender means that an established UI already exist to correct / remove / reject problematic edits. The multi-hypothesis nature of the edit generator also allows multiple possibilities to be generated, increasing the chances that there is one that is both usable and unproblematic.

\section{Limitations} \label{sec:limitations}

\paragraph{\textbf{Cost and speed of inference }}
Our system uses state of the art vision-language models. We've demonstrated this with GPT-4V, which as of March of 2024, remains very expensive and high-latency. For each of the material examples shown in the paper synthesized by a $4 \times 8$ dimensional system, the cost stands at just under \$3 per material, the  bulk of which is spent on the edit generator $G$. In practice, it's likely that BlenderAlchemy will need to synthesize many candidate materials for one to be usable for the end product, further increasing the average costs per \textit{usable} material. Beyond optimizing our system further (e.g. instead of pairwise comparisons for the visual state evaluator, ask it to choose the best candidate in batches),  we expect that the cost and speed of VLMs will substantially improve in the near future. Efforts that develop applications that can run large pretrained models locally \cite{ollama} also hold promise that we can further lower the latency/costs by running open-source pretrained models locally, concurrently with 3D design processes.

\paragraph{\textbf{Library of skills}}
We do not develop a library of skills that our procedure can use, which have shown to be important in \cite{wang2023voyager,de2023llmr,sun20233dgpt}. Such libraries are likely to be extremely domain-specific (library tools used by a material editor would be very different than animation), and will be the subject of our future work.

\paragraph{\textbf{Edit-only}}
We've made \textit{editing} 3D graphics our main objective, and though the method can, in theory, be trivially extended to iteratively edit an initial empty scene into the full generated scene, we do not demonstrate this. We believe that stronger VLMs than what exists today will be necessary to do end-to-end generation without humans in the loop, unless such generation is done in simple settings like \cite{yamada2024l3go} and \cite{sun20233dgpt}.


\paragraph{\textbf{VLM not fine-tuned to Blender scripting}}
GPT-4V still hallucinates when producing Blender python code with imports of libraries that don't exist or assigning values of  the wrong dimensions to fields. In such cases, our system does rejection sampling of edit generations based on whether errors are returned when run in Blender, but such a method can incur penalties in runtime and cost. Future work can look to fine-tune on real and synthetic datasets of Blender scripts and resultant visual outputs. We believe in such cases, the iterative refinement procedure of BlenderAlchemy may likely still be useful due to the inherent multi-step, trial-and-error nature of human design process in pursuit of a vague intent specification.

%
%

\end{document}